\pdfoutput=1

\documentclass[11pt]{article}

\usepackage[final]{acl}

\usepackage{times}
\usepackage{latexsym}

\usepackage[T1]{fontenc}

\usepackage[utf8]{inputenc}

\usepackage{microtype}

\usepackage{amsmath}
\usepackage{inconsolata}
\usepackage{adjustbox}
\usepackage{colortbl}
\definecolor{mygray}{gray}{.9}
\usepackage{makecell}

\usepackage{array}
\usepackage[symbol]{footmisc}

\usepackage{graphicx}

\title{ACE: A LLM-based Negotiation Coaching System}

\author{Ryan Shea\thanks{~~denotes equal contribution.},~
  Aymen Kallala$^{*}$,~
  Xin Lucy Liu$^{*}$,~
  Michael W. Morris\thanks{~~denotes equal advising.},~
  Zhou Yu\textsuperscript{\textdagger}\\
  Columbia University, New York, NY \\
  \texttt{rs4235@columbia.edu} \\
  \texttt{\{ak5078, xl2855, mwm82, zy2461\}@columbia.edu}
}

\begin{document}
\maketitle
\begin{abstract}
The growing prominence of LLMs has led to an increase in the development of AI tutoring systems. These systems are crucial in providing underrepresented populations with improved access to valuable education. One important area of education that is unavailable to many learners is strategic bargaining related to negotiation. To address this, we develop a LLM-based \textbf{A}ssistant for \textbf{C}oaching n\textbf{E}gotiation (ACE). ACE not only serves as a negotiation partner for users but also provides them with targeted feedback for improvement. To build our system, we collect a dataset of negotiation transcripts between MBA students. These transcripts come from trained negotiators and emulate realistic bargaining scenarios. We use the dataset, along with expert consultations, to design an annotation scheme for detecting negotiation mistakes. ACE employs this scheme to identify mistakes and provide targeted feedback to users. To test the effectiveness of ACE-generated feedback, we conducted a user experiment with two consecutive trials of negotiation and found that it improves negotiation performances significantly compared to a system that doesn't provide feedback and one which uses an alternative method of providing feedback.
\end{abstract}

\section{Introduction}

The rapid progress of LLMs in recent years has spurred the creation of more sophisticated AI tutoring systems \cite{sonkar-etal-2023-class}. These systems give learners easier access to training in areas outside what is typically available for most learners \cite{liang-etal-2023-chatback}. One area that falls into this category are social competencies, such as negotiation tactics. Negotiation skills are crucial as they help individuals maximize their gains in competitive situations across different areas of their professional and personal life. However, these skills are hard to learn from a traditional classroom lecture because they involve reflexive behavioral habits. They are typically taught through small seminars centered on role-playing exercises and instructor coaching, which are expensive and limited in access.

Therefore, effective negotiation training is not available to many populations who lack it yet need it the most, such as women and minorities. These groups particularly lack negotiation skills and are less accustomed to advocating for themselves \cite{BabcockLaschever_2003} which is a significant factor contributing to their relatively lower starting salaries and fewer promotion opportunities \cite{lu-and-zhao-2023-dei}. As such, increased coaching is needed to address gender and ethnic differences in negotiation performance \cite{amanatullah-2010-gender}. Recent research also finds that linguistic assertiveness skills, such as those used in negotiations, vary as a function of cultural/ethnic backgrounds, partly account for ethnic differences in promotions to leadership positions, and are amenable to training interventions \cite{lu-et-al-2020-dei, lu-et-al-2022-dei}. 

To democratize access to high-quality negotiation coaching, we proposed a LLM-based \textbf{A}ssistant for \textbf{C}oaching n\textbf{E}gotiation (ACE). ACE is designed to serve as a negotiation partner that provides learners with targeted feedback similar to what they would receive from an instructor in a seminar setting. To provide this feedback, we begin by collecting a dataset of negotiation transcripts between students in an Master's of business administration (MBA) negotiation class. The scenarios in these negotiations come from a rigorous curriculum and are designed to mirror real-world negotiation settings. Furthermore, the participants in these negotiations have been trained in bargaining tactics, unlike previous datasets which rely on crowd-workers \cite{lewis-etal-2017-deal, he-etal-2018-decoupling}.

Using this dataset along with expert consultations, we develop an annotation scheme to identify and correct mistakes that users make during their negotiations. Our annotations mirror actual error categories that instructors look for in negotiation classes. We then build ACE according to this scheme. ACE uses the annotation categories to identify users' mistakes and then provides targeted feedback based on the error definitions, along with in-context examples of feedback written by experts. Our system also includes a prompt-based negotiation chatbot agent which serves as a simulated negotiation partner for learners.

To test the efficacy of ACE, particularly the feedback it provides, we recruited a group of 374 users who participated in two trials of negotiation. We found that ACE significantly improves learning outcomes compared to users who negotiate with no coaching from our system. Our contributions are summarized as follows:

\begin{itemize}
\item We propose a novel negotiation coaching system called ACE, which utilizes LLMs to provide targeted, individualized feedback to users.
\item We develop a negotiation strategy annotation scheme to identify and correct user errors. Our annotation scheme is based on expert input and mirrors the kind of mistakes negotiation instructors look for in a seminar setting.
\item We release an annotated dataset of spoken negotiation transcripts between MBA students. The negotiation scenarios in our dataset are based on standardized business school curricula and are carefully designed to mimic real world settings.
\item We conduct an efficacy experiment with two trials of negotiations, finding that ACE boosted subjective and objective measures of tactical learning relative to a control system which does not provide feedback and an alternative method of providing negotiation feedback.
\end{itemize}

\section{Related Work}

\subsection{AI Tutoring Systems}
Generating automated training systems has been a long-standing issue for researchers in AI for education \cite{keuning-etal-2018-automatedfeedback}. One common application of such systems is the area of computing education \cite{koutcheme2024open}. Recent work has explored the use of LLMs to generate automatic feedback for students on programming assignments \cite{pankiewicz2023large} or for creating coding exercises \cite{denny-etal-2024-promptproblems}. These works have shown that state-of-the-art LLMs such as GPT-4 \cite{openai2023gpt4} can provide effective feedback to users and can achieve up to 80\% agreement with humans in certain scenarios \cite{zheng2023judging}.

Another application of these systems has been the field of language learning. These systems typically involve a chat agent which guides users through a conversation on a specific topic followed by feedback on the users' conversation \cite{qian-etal-2023-constrained-decoding, yuli-etal-2022-edubot}. These systems have been judged as helpful by English languages learners and have also had a demonstrable effect on actual learning outcomes \cite{liang-etal-2023-chatback}.

Prior work on developing systems for negotiation training has been fairly limited. These systems can only be interacted with by selecting from a list of pre-written options and deliver ``canned'' responses as a reply. Despite these limitations, prior work has shown that interacting with virtual agents can improve learners’ understanding of negotiations \cite{gratch-etal-2016-virtual}. 

\subsection{Negotiation}

Methods that have applied LLMs to the area of negotiation have been focused on building negotiation chat agents. These methods typically use existing negotiation data to perform supervised learning or offline reinforcement learning on a negotiation model \cite{lewis-etal-2017-deal,he-etal-2018-decoupling,verma-etal-2022-chai, zhan-etal-2024-lets}. More recent work has focused on examining and enhancing the negotiation capabilities of prompt-based negotiation agents \cite{ schneider2023negotiating, fu2023improving, zhan-etal-2024-lets}. 

Previous work on bargaining has focused on building chatbots to serve as negotiation partners. There is little work on providing user feedback. However, negotiation research has found that people do not learn negotiation tactics simply from the experience of bargaining; they need structured feedback and instruction \cite{Loewenstein-2006-learning-to-nego}. ACE is the first system to provide quality feedback using rigorous business school curricula to improve learning outcomes of negotiation.

\section{Background}
\label{sec:setting}
Negotiation is a general task and there are many ways to formulate a bargaining problem. In this section we give an overview of the the types of negotiation problems we consider when designing our coaching system and annotation scheme.

The negotiation settings we consider here are ones which involve one agent selling a single item to another agent. This type of negotiation is known as single-issue distributive bargaining \cite{lewicki-essentials}. The agent selling the item is referred to as the ``seller'' and the agent buying the item is called the ``buyer''. Both the buyer and seller have access to a ``role'' (also referred to as a ``negotiation scenario'') which provides private information about their options and preferences and public information about details of the object for sale and the range of market prices for this item. In a role-play simulation, participants use their role to prepare a strategy, this includes their upper limit or ``walk-away'' price, their target price, and the opening price that they will mention.  An example of buyer and seller roles for a used car negotiation can be found in Appendix \ref{sec:full_scenarios}.

\section{Dataset}
\label{sec:dataset}

Our dataset was collected in collaboration with an instructor who teaches a course on negotiation to MBA students. During this course, 50 students were randomly assigned to dyads for a negotiation task. They conducted three successive negotiations against three different counterparts, involving different kinds of cars. Before negotiating, students answered standard preparation questions and then began the negotiation while recording their conversation audio. We transcribed the conversation audio using OpenAI's Whisper API \cite{radford2022whisper} along with manual edits to construct our dataset.

A total of 42 dialogues were collected from this process, of which 40 were usable. A summary of the dataset statistics is in Table \ref{tab:data_stats}. We show the statistics for the entire dataset along with a breakdown by the negotiation task/scenario. The scenarios are based on the type of car being negotiated over in the exercise. We focused on negotiation over a Honda Accord (Task 1) as the transparent market range for such a product makes it typical of negotiation over a commodity item, see Appendix \ref{sec:full_scenarios}.

\begin{table}
\centering
\scalebox{0.7}{\begin{tabular}{p{0.63\linewidth} p{0.14\linewidth} p{0.14\linewidth} p{0.14\linewidth} p{0.1\linewidth}}
\hline
& \textbf{Task 1} & \textbf{Task 2} & \textbf{Task 3} & \textbf{Total} \\
\hline
\# of conversations & 14& 13&13&40 \\
Avg. \# of turns per conversation & 23.8& 22.3& 15.6& 20.6 \\
Avg. \# of tokens per turn & 31.4& 35& 37& 34.1 \\
Vocabulary size & 950& 1022& 820& 1723 \\
Deal \% & 93\%	& 31\%	& 100\%	& 75\% \\
Deal Amount & \$12.9k & \$7.3k & \$1.3k & \$7.2k \\
\hline
\end{tabular}}
\caption{A summary of statistics from our negotiation dataset.}
\label{tab:data_stats}
\end{table}

\begin{table*}[h]
\small
\begin{adjustbox}{width=\textwidth}
\begin{tabular}{ l| m{ 13.5cm} |m{ 3cm}}
\hline
\rowcolor{mygray}
\bf Role & \bf Utterance & \bf Annotation \\

\hline
\makecell{\\ Buyer}   &   \makecell{Hi, I'm new to California and I'm looking for probably a Honda Accord with reasonable mileage \\ around maybe \$11,000 to \$12,000. Do you have anything like that?}     &  \makecell{\textcolor{red}{Breaking the ice}\\ \textcolor{teal}{Giving the first offer}\\ \textcolor{teal}{Ambitious opening point}\\ \textcolor{red}{Including rationale}} \\\rowcolor{mygray}

Seller   &  Nice. We have something similar. We have a nice 2013 Honda. It does have a little bit more miles than that. It has about 50,000. It doesn't have any rust and it's in great condition. What's the price range you're looking to come out with?      &  \\

Buyer   &   Probably around \$11,000 or \$12,000.  & \makecell{\textcolor{teal}{Strong counteroffer} \\ \textcolor{red}{Including rationale}} \\\rowcolor{mygray}

Seller   &  Ooh, that's kind of rough. Our sticker price for this car is closer to \$14,000.       &   \\

Buyer   &  Ooh, yeah, that's definitely a little bit too much. Could I take it for a test drive maybe?      &    \\\rowcolor{mygray}

Seller   &  Sure.         &   \\

Buyer   &  Okay, great. Yeah, it's pretty good. What do you think about maybe \$12,500 and I would buy it today?  &  \makecell{\textcolor{teal}{Strong counteroffer} \\ \textcolor{red}{Including rationale}} \\\rowcolor{mygray}

Seller   &   \$12,500. I mean, could we call it even \$13,000?       &   \\

Buyer   &  Yeah, I could probably do \$13,000. &   \makecell{\textcolor{red}{Strong counteroffer} \\ \textcolor{red}{Including rationale}} \\\rowcolor{mygray}

Seller   &  All right.      &   \\

Buyer   &  All right.       &  \textcolor{red}{Strategic closing} \\\rowcolor{mygray}

Seller   &   Sounds great.      &  \\\hline

\end{tabular}
\end{adjustbox}
\caption{\label{tab:annotation_examples}An example negotiation dialogue from our dataset. Annotations in red indicate that the category was labeled as \textcolor{red}{False}. Annotations in green indicate that the category was labeled as \textcolor{teal}{True}.}
\end{table*}

\section{Annotation Scheme}
\label{sec:annotation_scheme}

We designed a negotiation error annotation scheme to identify and correct user mistakes. Our scheme is based on the dataset collected in Section \ref{sec:dataset}, expert input, and common distributive bargaining tactics from \citealp{lewicki-essentials}. 

We identified eight error categories that can be divided into \textbf{preparation errors} and \textbf{negotiation errors}. All the categories in our scheme are binary True/False labels, where a label of False indicates that a mistake has been made by the negotiator. Our annotation scheme is designed from the perspective of the buyer but can be easily adjusted to fit the seller's perspective.

The two categories below belong to \textbf{preparation errors}, which are mistakes in users' answers to a set of standard preparation questions before the negotiation. These categories are designed to identify whether the user prepared their negotiation strategy correctly.

\noindent
\textbf{1. Strategic walk-away point} assesses whether the user has properly analyzed the facts of the negotiation scenario to set an appropriate walk-away price. From the buyer's perspective, a walk-away price is the maximum amount they would pay to purchase the item in the negotiation. If the scenario outlines an explicit budget limit, then a strategic walk-away point is one which exactly matches the budget amount. Otherwise, we consider any point below the maximum market price to be strategic.

\noindent
\textbf{2. Strategic target price} evaluates the target price that the buyer sets before the negotiation. The  buyer's target price is strategic if it falls within the first third of the range between the minimum market price and the buyer's walk-away point. If the buyer sets their target below this range, it is too ambitious to be a realistic outcome to aim for. Conversely, if their target is above this range, it is too weak to test how far their opponent can be pushed in the negotiation.

The six categories below refer to \textbf{negotiation errors}, which are mistakes that users make ``at the bargaining table,'' during the negotiation dialogue.
\\
\noindent
\textbf{3. Breaking the ice} refers to whether or not the user began the negotiation with some social bonding.  A negotiator should spend their first conversational turn on social remarks unrelated to the negotiation issues.

\noindent
\textbf{4. Giving the first offer} indicates whether the user stated the first price offer in the conversation. Negotiators are advised to state their opening price first to anchor the negotiation in a favorable position \cite{lewicki-essentials}.

\noindent
\textbf{5. Ambitious opening point} assesses the tactical quality of the user's opening offer relative to their target price.
When the buyer proposes a price first, we consider the offer $O_1$ strong if: 
\begin{equation*}
   O_1 \leq 0.9*T
\end{equation*}

\noindent
with $T$ being the buyer's target price.
Otherwise, when the seller previously made an offer $S$, we consider the buyer's first offer strong if it creates a midpoint at or below their target price:
\begin{equation*}
    \frac{S+O_1}{2} \leq T
\end{equation*}

\noindent
\textbf{6. Strong counteroffer} assesses the quality of the user's first three proposals following their first offer. 
A counter-offer $O_t$ is considered strong if it's below the midpoint of the remaining bargaining range: 
\begin{equation*}
    O_t < \frac{O_{t-1} + min(S,W)}{2}
\end{equation*}

\noindent
where $O_{t-1}$ is the buyer's previous offer, $S$ is the seller's current offer, and $W$ is the buyer's walk-away point.

\noindent
\textbf{7. Including rationale} indicates whether the user's first four price offers were accompanied by a rationale. We define a rationale as any reasoning that supports a price offer \cite{lewicki-essentials}.

\noindent
\textbf{8. Strategic closing behavior} refers to whether the user closed the deal in ways that heighten the counterpart's commitment. The final two turns of the negotiation should contain either an acknowledgment of the opponent's negotiation skill or a recounting of their own concessions. The closing turns should not contain any celebratory statements about the negotiation outcome or any statements implying that the user got a better deal.

\begin{table}
\centering
\scalebox{0.87}{\begin{tabular}{p{0.485\linewidth} p{0.27\linewidth} p{0.23\linewidth}}
\hline
\textbf{Negotiation Error \qquad Category} & \textbf{Number of Turns with Errors} & \textbf{Number of Applicable Turns}\\
\hline
Breaking the ice & 28 & 40 \\
Giving the first offer & 15 & 40\\
Ambitious opening point & 18& 40\\ 
Strong counteroffer & 34 & 73\\
Including rationale & 25& 112\\
Strategic closing & 36& 40\\
\hline
\textbf{Preparation Error \qquad Category} & \textbf{Number of Errors} & \textbf{Number of Dialogues}\\
\hline
Strategic walk-away & 7 & 40 \\
Strategic target price & 15 & 40 \\
\hline
\end{tabular}}
\caption{A summary of our annotated negotiation dataset. Note that mistakes related to \textbf{negotiation errors} tend to be more common that \textbf{preparation errors}}
\label{tab:annotation_stats}
\end{table}

We annotated our collected dataset (Section \ref{sec:dataset}) based on this annotation scheme with an inter-annotator agreement of 0.87, according to Cohen's kappa \cite{cohens_kappa}. Inter-annotator agreement was calculated based on a subset of 288 dialogue turns annotated by two authors. A breakdown of the errors that buyers made according to the annotation scheme can be seen in Table \ref{tab:annotation_stats}. An example of an annotated conversation is given in Table \ref{tab:annotation_examples}. The most prevalent errors that buyers made were those in the \textbf{strategic closing} category, with 36 out of the 40 conversations containing an error related to this. On the other hand, we see that buyers did a good job including a rationale in their price offers, with only 25 of these errors present out of a total of 112 relevant conversation turns.

\section{Approach}
\label{sec:approach}

In this section, we outline our approach to designing ACE. We first provide a high-level overview of the system and how users progress through it. We then describe the components of the system, including the negotiation chatbot agent and feedback modules.

\subsection{System Overview}
\label{sec:sys_overview}
ACE begins by presenting users with a negotiation scenario similar to those in our dataset (Section \ref{sec:dataset}). This is followed by a set of negotiation preparation questions which ask users for their target price, walk-away price, and planned opening point. Users then proceed into a simulated negotiation with our negotiation agent until they reach an agreement. After that we provide users with the feedback associated with their preparation questions, referred to as ``preparation feedback.'' This is followed by ``negotiation feedback'' on the negotiators' linguistic performance, including both turn-specific points about their tactics and holistic points about their diction, tone, and politeness.

\subsection{Negotiation Agent}
\label{sec:chat_agent}
Our negotiation agent is based on a prompted version of GPT-4 \cite{openai2023gpt4}. Prior work has found that LLMs can achieve successful negotiation outcomes with proper prompting \cite{fu2023improving}. Therefore, we adopted this approach for our agent. We found that prompting GPT-4 with a full negotiation scenario (Appendix \ref{sec:full_scenarios}) resulted in nonstrategic oversharing; therefore, we use a summarized version as our instructional prompt (Table \ref{tab:chatbot_instructional_prompt}). 

Even with the summarized prompt, the agent is prone to making ``weak'' counteroffers, meaning that is often pushed to offering its walk-away price too easily. To avoid a ``pushover'' negotiation agent, we employ dynamic prompting. Specifically, we give the agent an initial ``reservation price'' in the instructional prompt, which is higher than the actual reservation price given in the negotiation scenario. We refer to this price point as the agent's ``subjective limit.'' We adjust the subjective limit as the conversation progresses and eventually set it to the true reservation price after several conversation turns have passed. This allows us to control the bot's counteroffers to a certain degree while allowing for variation in the offers presented. We initially set the bot's subjective limit to a random value within the range given by \textbf{strategic target price} (Section \ref{sec:annotation_scheme}). After the first turn, we update it to a price that corresponds to a \textbf{strong counteroffer}. This ensures that the bot gives robust counteroffers and will not reach its true reservation price too quickly.

\subsection{Preparation Feedback}
The preparation feedback we provide users is based on their answers to pre-negotiation preparation questions. We specifically look for errors corresponding to \textbf{strategic walk-away point}, \textbf{strategic target price}, and \textbf{giving the first offer} in our annotation scheme (Section \ref{sec:annotation_scheme}). Since each of these errors is identified with a mathematical formula, flagging them is trivial. We give feedback to users using either a hard-coded message with the correct answer and an explanation, or we prompt GPT-4o to generate feedback given the user's answer, the correct answer, and in-context examples of quality feedback written by expert negotiators (Appendix \ref{sec:prompts}). A full example of preparation feedback can be found in Appendix \ref{sec:full_feedback}.

\begin{figure*}[h]
    \centering
    \includegraphics[width=\textwidth]{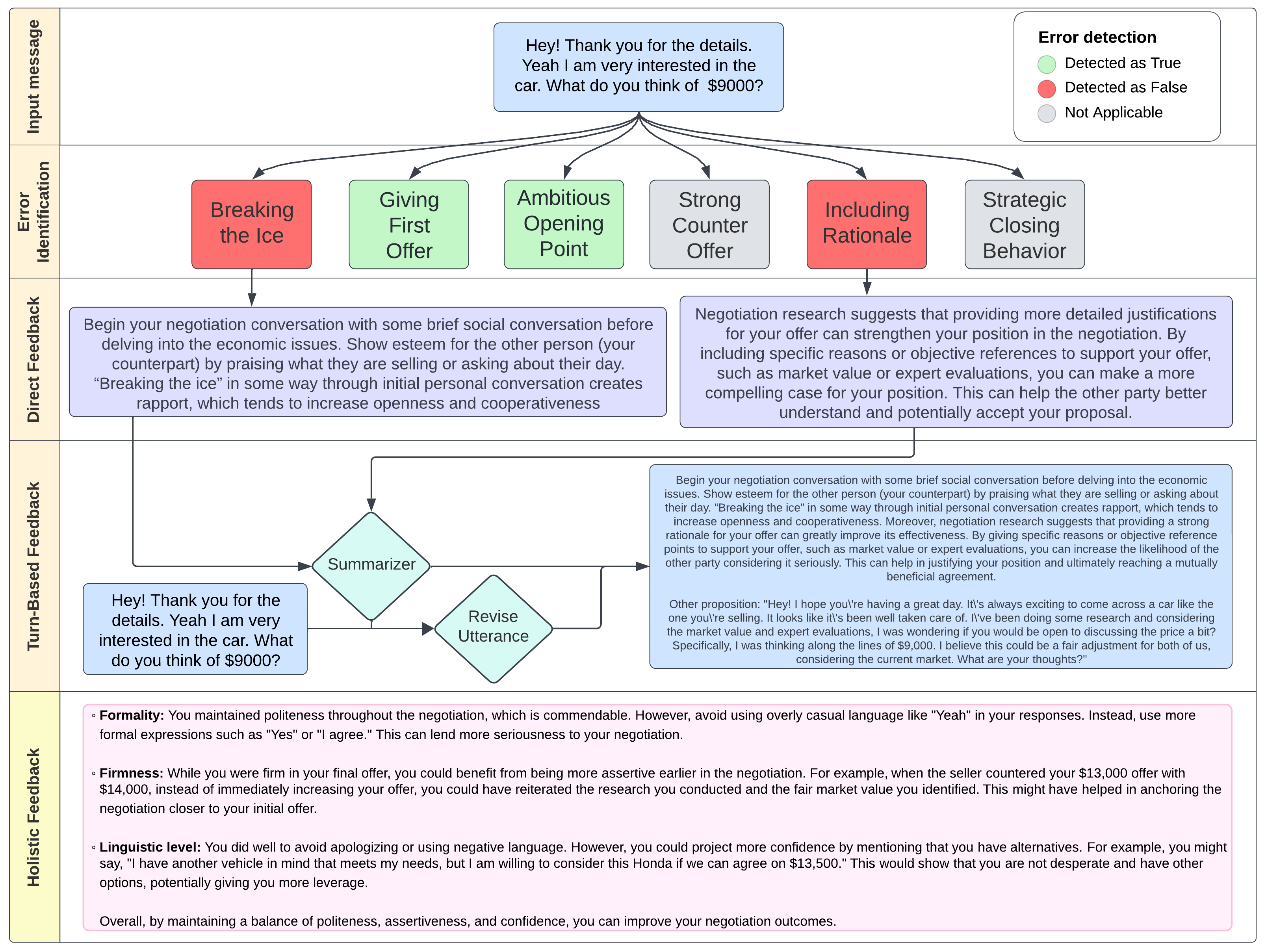}
    \caption{Diagram illustrating the turn-based feedback flow for ACE as well as an example of holistic feedback.}
    \label{fig:turn-based_feedback}
\end{figure*}

\subsection{Negotiation Feedback}
Negotiation feedback is provided based on users' conversations with the automated bargaining agent. We give users two categories of negotiation feedback. The first is feedback associated with individual conversation turns, which we call ``turn-based feedback''. The second is ``holistic feedback'' which corresponds to the conversation as a whole such as the conversational tone or politeness level.

\subsubsection{Turn-Based Feedback}
Figure \ref{fig:turn-based_feedback} gives an overview of the pipeline ACE uses to provide turn-based feedback. Our pipeline consists of three components: \textbf{1) error identification}, \textbf{2) direct feedback} and \textbf{3) utterance revision}. We describe each of these components in detail below.

\noindent
\textbf{Error identification.} To provide effective feedback, we start by detecting errors the user made at each conversational turn of their transcript. The errors we flag for turn-based feedback are based on the \textbf{negotiation errors} in our annotation scheme (Section \ref{sec:annotation_scheme}). We divide these errors into two groups and use a different strategy for detecting each of them. 

The first group of errors is based on price offers and involves applying a formula. These include categories such as \textbf{strong counteroffer} or \textbf{ambitious opening point}. The main challenge in identifying these mistakes is extracting the relevant price from the user utterance. We do this by prompting GPT-4 to extract the price. Our prompt consists of the user utterance followed by nine hand-written, in-context examples of successful price extractions (Table \ref{tab:price_extraction_prompt}). After extracting the price, we apply the relevant formula to detect errors. 

The second group of errors is based on the users' language rather than prices, such as \textbf{including rationale} or \textbf{breaking the ice}. For each turn, we use a set of classifiers, one for each of the three relevant categories, to determine whether the turn contains an error. We create our classifiers by prompting GPT-4o to output a True/False label for the turn. Our prompt consists of the user utterance along with the error definition. For more difficult categories, such as \textbf{including rationale}, the prompt also contains in-context examples from our collected dataset (Table \ref{tab:rationale_prompt}).

\noindent
\textbf{Direct feedback.} When we identify a turn with a mistake, we prompt our GPT-4o to give direct comments on each error committed. This includes an explanation of the error and the tactical value of correcting it, see Figure \ref{fig:turn-based_feedback} for examples. Our prompt consists of the conversation context, a definition of the error committed (Section \ref{sec:annotation_scheme} and Table \ref{tab:direct_feedback_prompts}), and one or two in-context examples of feedback written by negotiation instructors. We generate feedback for each mistake committed, which can result in the comments being quite lengthy. Therefore, in cases where the user commits more than one mistake in a turn, we give a summary of the error explanations as the final direct feedback.

\noindent
\textbf{Utterance revision.} Along with the direct feedback, we also present users with a revised version of their utterance with their mistakes corrected. We prompt GPT-4o with the user utterance and the direct feedback to generate the corrected utterance. Our prompt includes the user utterance, the direct feedback, and three in-context examples of hand-written utterance revisions (Table \ref{tab:re-edit_message_prompt}). 

The final turn-based feedback given to the user is the direct feedback concatenated with the revised user utterance. See Figure \ref{fig:turn-based_feedback} and Appendix \ref{sec:full_feedback} for complete examples.

\subsubsection{Holistic Feedback}
The holistic feedback presented to the user is not based on any specific errors identified within the transcript. Instead, we prompt GPT-4o to comment on the linguistic aspects of the user's conversation. The aspects we focus on are formality, firmness, and linguistic level. For formality, the user should stay polite and avoid being rude or pushy. It's also better for users to be firm and assertive in their language, as studies have shown that this communication style leads to better deals \cite{jeong-etal-2019-negotiation-firmness}. In terms of linguistic level, users should avoid apologizing or using language that could be interpreted as a personal attack \cite{FisherUryPatton11-getting-to-yes}. We prompt the model with a summary of the attributes for these three aspects and have it generate feedback. We have the model quote specific phrases from the users' transcript to make the comments more targeted and personalized (Table \ref{tab:overall_feedback_prompt}). See Figure \ref{fig:turn-based_feedback} for a full example.

\section{Evaluation}
\label{sec:evaluation}

\subsection{Error Identification Evaluation}
We use a subset of the annotated data from Section \ref{sec:annotation_scheme} to evaluate how well ACE can identify mistakes. This subset consists of 26 dialogues (with a total of 494 conversational turns) which were excluded during the creation and testing of our coaching system. Since the \textbf{preparation errors} in our annotation scheme are trivial to identify, we measure how well ACE can classify \textbf{negotiation errors}.

Table \ref{tab:automatic_eval} shows a breakdown of the system's accuracy. Using human annotations as our ground truth labels, we measure how well ACE predicts these labels. The system can identify mistakes with a high accuracy of at least 0.90 for all error categories. However, it performs worse in terms of precision and recall, with the F1 score ranging from a low of 0.54 for \textbf{strategic closing} to a high of 0.97 for the \textbf{giving the first offer} category. This suggests that ACE has difficulty balancing precision and recall for some annotation categories. Low recall in particular seems to be driving the lower scores, which implies that ACE has trouble identifying true positive cases for some difficult categories.

\begin{table}
\centering
\scalebox{0.75}{\begin{tabular}{p{0.39\linewidth} p{0.2\linewidth} p{0.19\linewidth} p{0.12\linewidth} p{0.19\linewidth}}
\hline
\textbf{Error Category} & \textbf{Accuracy} & \textbf{Precision} & \textbf{Recall} & \textbf{F1 Score} \\
\hline
Breaking the ice & 0.98	& 0.99 & 0.76 & 0.83 \\
First offer & 0.99 & 0.95 & 0.91 & 0.93 \\
Strong first offer & 0.98 & 0.91 & 0.83 & 0.85 \\
Strong counteroffer & 0.96 & 0.74 & 0.73 & 0.73 \\
Including rationale & 0.90 & 0.81 & 0.63 & 0.67 \\
Strategic closing & 0.94 & 0.72 & 0.53 & 0.54\\
\hline
\end{tabular}}
\caption{A table indicating how accurately ACE can identify user mistakes. Our system is able to detect errors with high accuracy, but performs worse in terms of precision and recall. This is reflected in the lower F1 scores  for \textbf{including rationale} and \textbf{strategic closing}.}
\label{tab:automatic_eval}
\end{table}

\subsection{ACE-generated Feedback Evaluation}
We next evaluate the feedback generated by ACE via a user experiment. Our experiment consists of two pilot studies and a main experiment. In our first pilot study (Pilot Study A) we recruited 100 native English-speaking U.S. participants from Prolific and asked them to perform two trials of used car negotiation. All participants in all trials were assigned the role of the buyer based on the used car scenario in Figure \ref{fig:honda_scenario_buyer}. Participants were randomly assigned to treatment or control conditions, and ACE feedback was only provided in the first trial of the treatment condition. Users in the control condition were given no feedback. The results of the pilot study showed that the improvements in the the treatment condition were significantly larger than in the control condition. Additional details for Pilot Study A can be found in Appendix \ref{sec:pilot_studies}.

\begin{table*}
\centering
\scalebox{0.83}{\begin{tabular}{p{0.282\linewidth} p{0.19\linewidth} p{0.165\linewidth} p{0.173\linewidth} p{0.273\linewidth}}
\hline
 & \textbf{ACE \qquad \qquad Condition (N=119)} & \textbf{Other-feedback (N=129)} & \textbf{No-feedback \qquad \qquad Baseline (N=126)} & \textit{Between-person Comparison} \\
\hline
\textbf{$\mathbf{1}\textsuperscript{st}$ Trial: Used Car} & \$12,891 (623.25) & \$12,889 (779.13) & \$12,948 (789.95) &	$F$ (2, 371) = 0.27, $p$ = 0.77 \\
\textbf{$\mathbf{2}\textsuperscript{nd}$ Trial: Summer Sublease} & \$7,528 (658.72) & \$7,751 (459.72) & 7,827 (431.65) & $F$ (2, 371) = 10.79, $p$ < 0.001 \\
\textit{Within-person Comparison} & $t$ = -2.97, $p$ = 0.003 & $t$ = 0.30, $p$ = 0.76 & $t$ = 1.03, $p$ = 0.30 &	$F$ (2, 371) = 8.80, $p$ < 0.001 \\
\hline
\end{tabular}}
\caption{\label{tab:human_eval}
A summary of outcomes from the human evaluation of ACE. Our results show that buyers who received feedback from ACE performed significantly better in a following negotiation than those who didn't. 
\textit{Note}: 1. Among 371 participants, 52\% were female; $\Bar{x}_{age}$ = 34.48 years, $SD_{age}$ = 8.97; 60.1\% were White/Caucasian; 54.4\% had a bachelor’s degree and above. The median completion time for this study was 39.68 minutes. 2. Standard deviations are presented in parentheses. 3. $d$ in reported results stands for Cohen's d,  a measure of effect size, calculated as the difference between two means divided by the pooled standard deviation of the data.
}
\end{table*}

To exclude the possibility that the improvements were due to the learning of situation-specific inert knowledge limited to the used car scenario, we developed a new negotiation scenario. This new scenario involves negotiating over a the price of summer sublease and can be seen in Figures \ref{fig:sublease_buyer} and \ref{fig:sublease_seller}. We tested its difficulty in Pilot Study B to ensure it has no difference with the used car scenario. We recruited 46 Prolific participants to negotiate against the ACE chat agent in the role of the buyer. T-tests revealed no significant differences between the standardized deal prices of the two scenarios (p = 1). Their distributions also showed no significant differences (Mann-Whitney U test: p = 0.9; Levene test: p = 0.6). Therefore, we can confidently state that the difficulty of the two scenarios is equivalent so that any differences that emerge across trials can not come from the scenario itself.

After the two pilot studies were completed we performed our main experiment. The main experiment was designed based on Pilot Study A but with several enhancements. The experimental details and results of this study are given in the following sections.

\subsubsection{Experimental Setup}
For our main experiment, we recruited 390 U.S.-based native English speakers from Prolific. This sample size was chosen based power analysis for a medium-size effect ($f=0.2$) which determined that $N > 390$ is required for robust results of ANOVA tests across three conditions. Each participant was paid \$8 for completing this 40 minute online study. We obtained a total of 374 valid responses. 

After consenting to this anonymous study, participants engaged in two trials of negotiation with a bot that simulated standard distributive bargaining tactics. All participants were assigned the role of the buyer. The first trial of negotiation was based on the used car scenario in Figure \ref{fig:honda_scenario_buyer} and the second trial was based the summer sublease scenario in Figure \ref{fig:sublease_buyer}. After each negotiation, they answered a round of questions. Participants were randomly assigned to one of three conditions. In the ACE condition, participants received ACE-generated feedback. In the ``Other-feedback'' condition, participants were given feedback based on the method described in \citealp{fu2023improving}. This approach leverages the zero-shot capabilities of GPT-4 to give feedback and has been shown to improve the negotiation performance of LLMs \cite{fu2023improving}. Lastly, participants in the ``No-feedback'' condition received no feedback on their performance. The procedure is illustrated in Figure \ref{fig:experiment_diagram}. We pre-registered the design for our user experiment at \href{https://aspredicted.org/NPR_36R}{aspredicted.org/NPR\_36R}.

Before starting, participants in all conditions answered four questions about their goals in the negotiation (e.g., how important it would be for them to reach a favorable deal, to be a tough bargainer). As shown in Table \ref{tab:eval_motivation}, participants in the conditions showed no differences in any of these goals, suggesting that the differences in learning outcomes between conditions should be attributed to the effectiveness of ACE rather than their motivation level. 

For the first trial, participants in the ACE condition ($N=119$) proceeded through the ACE system as described in Section \ref{sec:sys_overview}. Participants in the Other-feedback condition ($N=129$) proceeded through a similar system but were given feedback according to the method outlined in \citealp{fu2023improving}. Users in the No-feedback condition ($N=126$) followed a similar procedure but were not given any feedback at the end of the negotiation. In every condition, we asked participants a series of reflection questions to ensure they digested any feedback they were given and prepared their strategy for the next round of negotiation (See Table \ref{tab:ace_reflection}).

For the second trial, participants in both conditions practiced with a negotiation agent with no feedback provided, as participants in the No-feedback condition did in their first trial. Then we asked all participants their subjective perceptions of improvement in the second negotiation compared to the first. A sample item was “\textit{Compared to the first round of negotiation, in the second negotiation, I felt more confident.}” See details in Appendix \ref{sec:experiment_details}. 

\subsubsection{Results}

\noindent
\textbf{Objective Improvement.} We extracted the final deal price participants settled on from their chat history with the negotiation agent. Lower prices indicate more successful negotiations for the participants as buyers, so reaching a lower price in the second trial provides objective evidence of learning. As may be seen in Table \ref{tab:human_eval}, the performance improvement was significant in the ACE condition ($t = 2.97$, $95\% \text{ CI} = [0.14\text{ ,} 0.70]$, $p = 0.003$, $d = 0.38$), while no significant change in deal price was observed in Other-feedback condition or the No-feedback condition ($t = -0.23$, $p = 0.82$, $t = -1.03$, $p = 0.30$). A significant 2 $\times$ 3 two-way mixed ANOVA ($F(2, 368) = 8.67$, $p < 0.001$) indicated that the magnitude of improvement differed across conditions. 

Further, 2 $\times$ 2 ANOVAs suggested that the improvement in the ACE condition was significantly higher than in the No-feedback condition ($F(1, 246) = 12.82$, $p < 0.001$) or the Other-feedback condition ($F(1, 240) = 10.04$, $p = 0.002$), confirming the effectiveness of the ACE condition compared to the other conditions. Additionally, the lack of a significant difference between the Other-feedback and No-feedback condition ($F(1, 250) = 0.57$, $p = 0.451$) indicates that Other-feedback did not improve negotiation performance, underscoring the need for detailed, targeted feedback like provided by the ACE system. Additionally, this suggests that the kind of feedback that can help LLMs improve their task performance may not aid human performance. The feedback method in \citealp{fu2023improving} has been show to improve the abilities of LLM negotiators (Appendix \ref{sec:technical_approaches}) but our user experiment demonstrates that this type of coaching may not be useful for humans.


\noindent
\textbf{Subjective Improvement.} Participants in the ACE condition also reported a higher score for perceived improvement in the second negotiation ($\Bar{x} = 4.34$, $SD = 0.62$) compared to those in the Other-feedback ($\Bar{x} = 4.19$, $SD = 0.70$) and No Feedback ($\Bar{x} = 4.17$, $SD = 0.66$) conditions.  A significant 2 $\times$ 3 two-way mixed ANOVA ($F (2, 371) = 2.59$, $f = 0.12$, $p = 0.08$) indicated that the magnitude of improvement differed across conditions.

\section{Conclusion}
AI tutoring systems have the potential to democratize high-quality education in key areas such as negotiation. Prior work has shown that additional negotiation coaching is needed to correct for systematic gender and ethnic differences in bargaining performance \cite{amanatullah-2010-gender, lu-and-zhao-2023-dei}. To address this need, we built ACE to mimic the coaching learners would receive from a professional negotiation instructor in a seminar setting. ACE was built based on a dataset and annotation scheme created in collaboration with experienced negotiation instructors. We evaluated our system and confirmed its error identification accuracy and feedback effectiveness. Results from a user experiment demonstrated that users who interact with ACE improved their negotiation performance significantly compared to those who do not recieve ACE-coaching.

\section*{Limitations}
There are some limitations to the ACE system. The challenge for our negotiation agent is that LLM models tend to be agreeable and are not hardball bargainers. They respond to the user, so if users refused to discuss price, our negotiation agent wouldn't talk about price either. Additional work is needed to ensure our negotiation agent can guide users towards discussing price, especially when learners may be reluctant to fully engage in the negotiation. 

Another limitation of ACE is that it does not retain any memory of previous user interactions. As a result, the utility of the feedback and negotiation agent may diminish for individuals who want to engage in repeated interactions. More work is required to enable our system to retain previous user interactions and tailor the chat agent and feedback accordingly.

Finally, the annotation scheme and feedback method we built is based on an ``American'' style of negotiation. Other cultures have different standards for effective bargaining and what constitutes a mistake in negotiation. Therefore, ACE may not be as useful for individuals who want to improve their negotiation capabilities in other cultural contexts.

\section*{Acknowledgments}
We would like to thank Shehan Panditharatne, Xianglu Tang, and Aylin Hadzhieva for their help in implementing our user experiments.

\bibliography{custom}

\clearpage
\newpage

\appendix

\section{Additional Experimental Details}
\label{sec:experiment_details}
\subsection{Objective Performance}
In addition to the deal price they settled on, another indicator of objective performance in negotiation can be how long they persist before caving in. If participants have learned more negotiation tactics and know-hows, they should be able to persist longer in the second trial. Therefore, we also did ANOVA tests on the number of negotiation turns and the duration of their negotiations, and found a similar pattern to the objective results. That is, participants in the feedback group had more negotiation turns and longer negotiation duration in the second trial. Detailed analysis of the negotiation duration (in seconds) can be found in Table \ref{tab:eval_duration}.

\subsection{Post-Negotiation Reflection Questions}
Table \ref{tab:ace_reflection} shows the set of reflection questions we asked users after they completed the first negotiation in the experiment. These questions were asked in all conditions of the to ensure that no differences in negotiation ability emerged from users reflecting on their performance.

\begin{table*}
\centering
\scalebox{0.85}{\begin{tabular}{p{0.34\linewidth} p{0.19\linewidth} p{0.165\linewidth} p{0.173\linewidth} p{0.18\linewidth}}
\hline
\textbf{Survey Item} & \textbf{ACE \qquad \qquad Condition (N=119)} & \textbf{Other-feedback (N=129)} & \textbf{No-feedback \qquad \qquad Baseline (N=126)} & \textit{Between-person Comparison} \\
\hline
1. To reach a favorable deal & 4.47 (0.64) & 4.49 (0.66) & 4.51 (0.61) & $F$ = 0.13, $p$ = .88\\
2. To reach a fair deal & 4.34 (0.7) & 4.29 (0.79) & 4.46 (0.68) & $F$ = 1.63, $p$ = .20\\
3. To maintain an agreeable process & 4.14 (0.84) & 4.20 (0.64) & 4.12 (0.77) & $F$ = 0.38, $p$ = .69\\
4. To be a tough bargainer & 3.35 (0.93) & 3.36 (1.05) & 3.38 (1.00) & $F$ = 0.03, $p$ = .97\\
\hline
\end{tabular}}
\caption{\label{tab:eval_motivation}
Goal Set for Negotiations. Unless specified otherwise, all our questions use a five-point Likert scale. We asked participants “\textit{Before we start, please tell us about your approach to negotiations by rating the items below. It is important to me... (1 = Strongly disagree, 5 = Strongly Agree).}” The Cronbach's alpha of our 4-item measure is 0.90, indicating high internal reliability of the scale. No difference in any of these goals excludes motivation level as a predictor for the differences that emerged between the two conditions.}
\end{table*}

\begin{table*}
\centering
\scalebox{0.83}{\begin{tabular}{p{0.282\linewidth} p{0.19\linewidth} p{0.165\linewidth} p{0.173\linewidth} p{0.273\linewidth}}
\hline
 & \textbf{ACE \qquad \qquad Condition (N=119)} & \textbf{Other-feedback (N=129)} & \textbf{No-feedback \qquad \qquad Baseline (N=126)} & \textit{Between-person Comparison} \\
\hline
\textbf{$\mathbf{1}\textsuperscript{st}$ Trial: Used Car} & 418.20 
(307.16) & 427.65 (310.64) & 420.52 (278.41) &	$F$ (2, 371) = 0.03, $p$ = 0.97 \\
\textbf{$\mathbf{2}\textsuperscript{nd}$ Trial: Summer Sublease} & 503.50 (305.67) & 402.06 (238.43) & 362.61 (218.19) & $F$ (2, 371) = 9.92, $p$ < 0.001 \\
\textit{Within-person Comparison} & $t$ = -2.15, $p$ = 0.03 & $t$ = 0.73, $p$ = 0.46 & $t$ = 1.86, $p$ = 0.06 &	$F$ (2, 371) = 10.40, $p$ < 0.001 \\
\hline
\end{tabular}}
\caption{\label{tab:eval_duration}
Time duration of negotiations in seconds.
}
\end{table*}

\begin{table*}
\centering
\begin{tabular}{p{5.6in}}
\hline \textbf{Subjective Improvement Questions} \\ \hline
1. I felt more confident.\\
2. I felt more comfortable bargaining.\\
3. I expressed myself better.\\
4. I had a better understanding of the process.\\
\hline
\end{tabular}
\caption{Subjective improvement questions for participants. The first three items are adapted from an existing assertiveness scale \cite{Wallen2017}, and the last item is added to serve our study design. This scale's Cronbach Alpha is 0.90, which suggests its high reliability and excellent internal consistency among the items.
}
\label{tab:subjective_improvement}
\end{table*}

\begin{table*}
\centering
\begin{tabular}{p{5.6in}}
\hline \textbf{Post-Negotiation Reflection Questions} \\ \hline
1. Based on the feedback, what should be your walkaway point, your target point, and your opening point, respectively? \\
2. Based on the feedback, what can be compelling rationale for your offers and useful questions to elicit information or persuade the seller to make concessions? \\
3. What tips about your performance did you receive about the early phase of your negotiation conversation? Accordingly, what would you strive to do next time? \\
4. What tips about your performance did you receive about the later phase of you negotiation conversation? Accordingly, what would you strive to do next time? \\
\hline
\end{tabular}
\caption{Post negotiation reflection questions for participants.
}
\label{tab:ace_reflection}
\end{table*}

\section{Pilot Study A}
\label{sec:pilot_studies}

The procedure and results from Pilot Study A are described in this section. We began by recruiting 100 U.S.-based native English speakers from Prolific. Each participant was paid \$8 for completing this online study involving two trials of a car negotiation, each followed by survey questions. We obtained a total of 96 effective responses.

\begin{figure*}[h]
    \centering
    \includegraphics[width=\textwidth]{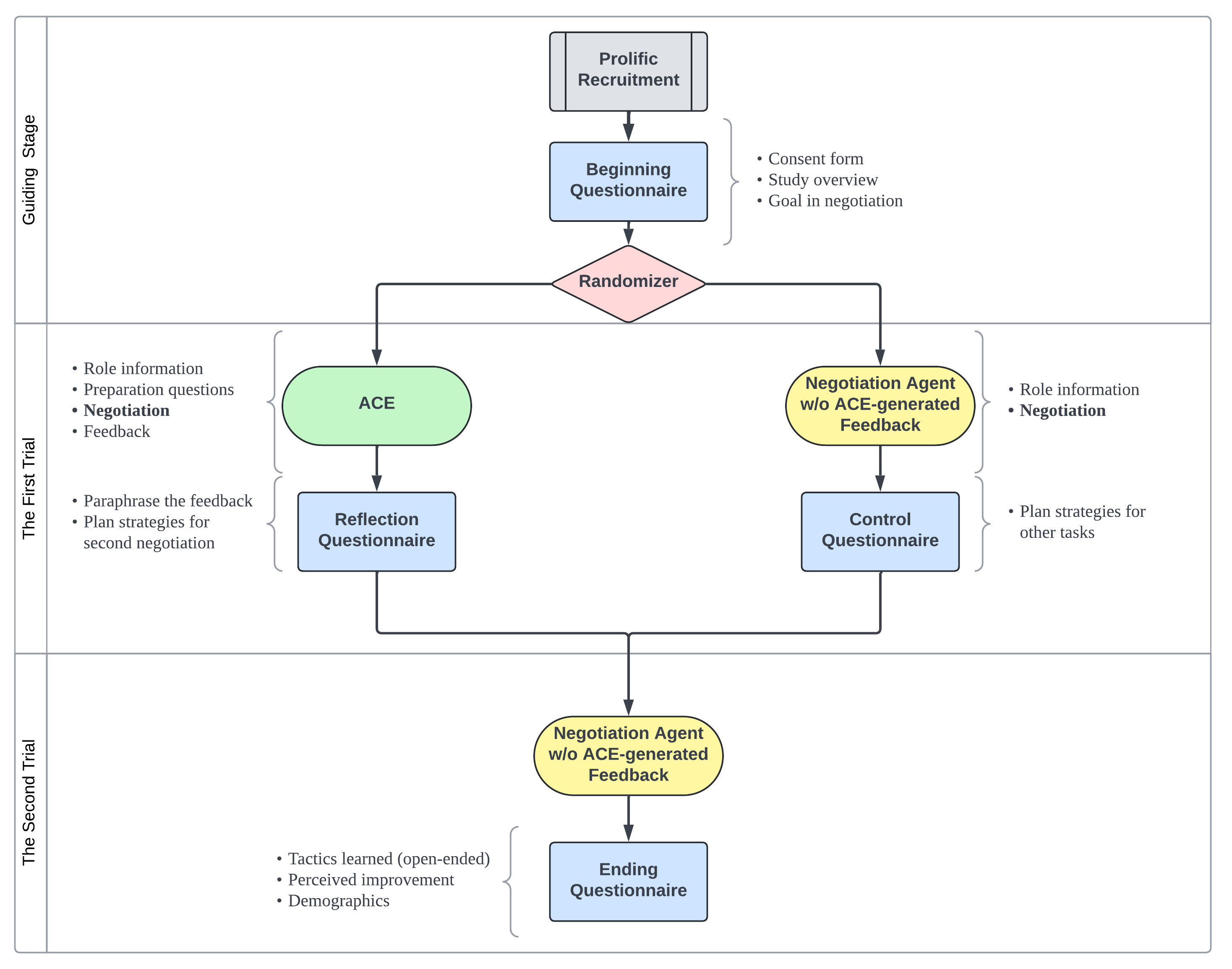}
    \caption{Experiment diagram for Pilot Study A, we designed our experiment in Qualtrics.}
    \label{fig:experiment_diagram_pilot_A}
\end{figure*}

After consenting to the study, participants engaged in two trials of a used car negotiation with the ACE chatbot. All participants were assigned the role of the buyer in both trials based on the scenario in Figure \ref{fig:honda_scenario_buyer}. Participants were randomly assigned to treatment or control conditions, and ACE coaching was only provided in the first trial of the treatment condition. The procedure is illustrated in Figure \ref{fig:experiment_diagram_pilot_A}. 



For the first trial, participants in the ACE condition ($N$ = 46) proceeded through the ACE system as described in Section \ref{sec:sys_overview}. To ensure they digest their feedback thoroughly, we asked them a series of questions to guide their reflection on the feedback (See Table \ref{tab:ace_reflection}). Participants in the Control condition ($N$ = 50) were not asked preparation questions nor given any feedback. To match the workload across conditions, they were asked a series of filler questions instead (See Table \ref{tab:control_reflection}).

For the second trial, participants in both conditions practiced with a negotiation agent with no ACE feedback provided, as participants in the control condition did in their first trial. Then we asked all participants their subjective perceptions of improvement in the second negotiation compared to the first.

\begin{table*}
\centering
\begin{tabular}{p{0.26\linewidth} p{0.18\linewidth} p{0.18\linewidth} p{0.25\linewidth}}
\hline
 & \textbf{ACE \qquad \qquad Condition (N=46)} & \textbf{Control \qquad \qquad Condition (N=50)} & \textit{Between-person \qquad Comparison} \\
\hline
\textbf{$\mathbf{1}\textsuperscript{st}$ Negotiation} & \$12,928 (693.11) & \$13,161 (429.08) &	$t$ = -1.96, $p$ = 0.054 \\
\textbf{$\mathbf{2}\textsuperscript{nd}$ Negotiation} & \$12,485 (967.77) & \$13,091 (621.69) & $t$ = -3.61, $p$ < 0.001 \\
\textit{Within-person Comparison} & $t$ = 2.52, $p$ = 0.014 & $t$ = 0.66, $p$ = 0.514 &	$F$ (1, 94) = 4.42, $p$ = 0.038 \\
\hline
\end{tabular}
\caption{\label{tab:human_eval_pilot}
A summary of outcomes from the human evaluation of ACE in Pilot Study A. Our results show that buyers who received feedback from ACE performed significantly better in a following negotiation than those who didn't. 
\textit{Note}: 1. Among 96 participants, 43\% were female; $\Bar{x}_{age}$ = 34.04 years, $SD_{age}$ = 9.14; 55\% were White/Caucasian; 51\% had a bachelor’s degree and above. The median completion time for this study was 36.29 minutes.
}
\end{table*}



\begin{table*}
\centering
\begin{tabular}{p{5.6in}}
\hline \textbf{Reflection Questions for the Control Condition in Pilot Study A} \\ \hline
1. If you want to develop a new hobby, what should be your first step? Please write down a tactical plan. \\
2. Can you think of any useful tactics to learn a new foreign language? \\
3. If you aim to improve your performance at work, what should you do? Please write down a tactical plan. \\
4. In applying to graduate school, what are some steps that a student can take to raise their GPA? \\
\hline
\end{tabular}
\caption{Filler questions given to users in the Control condition in Pilot Study A after they completed their first negotiation. These questions are given to match the workload between conditions.
}
\label{tab:control_reflection}
\end{table*}

\begin{table*}
\centering
\begin{tabular}
{p{0.25\linewidth} p{0.15\linewidth} p{0.15\linewidth} p{0.15\linewidth} p{0.15\linewidth}}
\hline
\textbf{Error Category} & \textbf{Accuracy} & \textbf{Precision} & \textbf{Recall} & \textbf{F1 Score} \\
\hline
Breaking the ice & 0.99	& 0.96 & 0.94 & 0.95 \\
First offer & 0.98 & 0.82	& 0.81 & 0.81 \\
Strong first offer & 0.97 & 0.68 & 0.67 & 0.65 \\
Strong counteroffer & 0.91 & 0.46 & 0.50 & 0.48 \\
Including rationale & 0.87 & 0.52 & 0.54 & 0.51 \\
Strategic closing & 0.96 & 0.68 & 0.78 & 0.66\\
\hline
\end{tabular}
\caption{A table indicating how accurately ACE can identify user mistakes when prompted with GPT-3.5-turbo-0125 instead of GPT-4o. Our system is still able to detect errors with high accuracy, but performs worse overall, especially with categories involving numbers such as for \textbf{strong counteroffer} and \textbf{strong first offer}.}
\label{tab:automatic_eval_gpt_3.5}
\end{table*}

\subsection{Results}

\noindent
\textbf{Objective Improvement.} We measured improvement by looking at the change in deal price accross negotiation trials. As may be seen in Table \ref{tab:human_eval}, the performance improvement was significant in the ACE condition but not the Control condition. A two-level within-person $\times$ two-level between-person factorial ANOVA results reveal a significant interaction effect. This indicates that participants learned more from the experience of negotiating against the bot when it was surrounded by ACE coaching.

\noindent
\textbf{Subjective Improvement.} Participants in the ACE condition also reported a higher score for perceived improvement in the second negotiation ($\Bar{x} = 4.23$, $SD = 0.72$) compared to those in the Control condition $(\Bar{x} = 3.93$, $SD = 0.80$; $t = 1.96$, $p = .053$). 



\subsection{Post-Negotiation Reflection Questions}
Tables \ref{tab:ace_reflection} and \ref{tab:control_reflection} show the set of reflection questions we asked users after they completed the first negotiation in the experiment. Users were given these instructions when answering the reflection questions: \textit{Please answer these questions in your own words. No AI-generated text is allowed. Use at least 30 characters for each of these open-ended questions.}

\section{Alternative Technical Approaches}
\label{sec:technical_approaches}

\subsection{Building ACE}
In this section we briefly go over some alternative technical approaches we explored to build the ACE system. We tested using smaller models for the error detection module such as GPT-3.5-turbo, T5, and Mistral. In general, we found that these models perform worse than GPT-4 and GPT-4o, especially when it comes to price extraction. This can be seen in Table \ref{tab:automatic_eval_gpt_3.5} in the decreased performance for “strong first offer” and “strong counter offer” when compared to GPT-4o. GPT-4 and GPT-4o performed comparably so we chose GPT-4o because it is faster.

As for the other components of our system, such as direct feedback and utterance revision, we do not have data and therefore could not perform automatic evaluations. Instead, we relied on small scale expert evaluations of outputs to choose the best models. For each model we tested (GPT-4, GPT-4o, and GPT-3.5) we generated 10-15 examples of output and presented them to two experts who then selected which output they preferred. Our evaluations showed that GPT-4 and GPT-4o performed the best and we ultimately chose GPT-4o due to its speed.

For our chat agent, we tested both GPT-4 and GPT-4o by having two experts perform 10 negotiations with the model and rate its performance. We found that GPT-4 was better based on this evaluation. Prior works have extensively tested the negotiation performance of various LLM chat agents and also found that GPT-4 achieves quality performance \cite{bianchi2024llmsnegotiatenegotiationarenaplatform}.

\subsection{Alternate Feedback Approaches}

In this section we provide some additional details about the alternative feedback approach in \citealp{fu2023improving}. As mentioned in Section \ref{sec:evaluation}, this method uses the zero-shot capabilities of GPT-4 to give negotiation feedback. The method was originally designed for improving LLM negotiation performance, the procedure goes as follows. First two LLMs participate in a negotiation, then another LLM provides three suggestions on how the buyer/seller can improve their performance in the next negotiation. The initial negotiation along with the suggestions are used as the prompt for the following negotiation. \citealp{fu2023improving} demonstrates that this feedback is effective at improving performance between LLM negotiators. However, the negotiation setting they tested is fairly basic. It involves a two parties negotiating over the sale of a balloon with the buyer's and seller's opening offers hard-coded at \$10 and \$20, respectively. Therefore we tested whether this approach remains effective when applied to our more complex setting of used car negotiation (Figures \ref{fig:honda_scenario_buyer} and \ref{fig:honda_scenario_seller}).

We conducted 200 simulated negotiations between a LLM buyer and seller. In 100 of the negotiations, no feedback was given to either party. In the other 100 negotiations the buyer received feedback according to \citealp{fu2023improving}. The average deal price with no feedback given was \$12,984 ($SD=554)$ and the average deal price with feedback given to the buyer was \$12,788 ($SD=749$). A two sample t-test shows that the buyer's improvement was statistically significant $p=.038$. This demonstrates that the feedback method in \citealp{fu2023improving} is able to enhance the negotiation performance of LLMs in more complex settings such as used car negotiation. This also illustrates that the lack of improvement of human negotiators when given feedback based on this method (Section \ref{sec:evaluation}) is not due to the nature or complexity of our scenario. Instead the lack of improvement likely stems from inherent differences in what constitutes effective feedback for LLMs versus humans. This further emphasizes the need for extensive human testing when building educational systems as LLMs may not serve as a realistic proxy for human learners.

\section{Negotiation Scenarios}
\label{sec:full_scenarios}

\begin{figure*}[h]
    \centering
    \includegraphics[width=\textwidth]{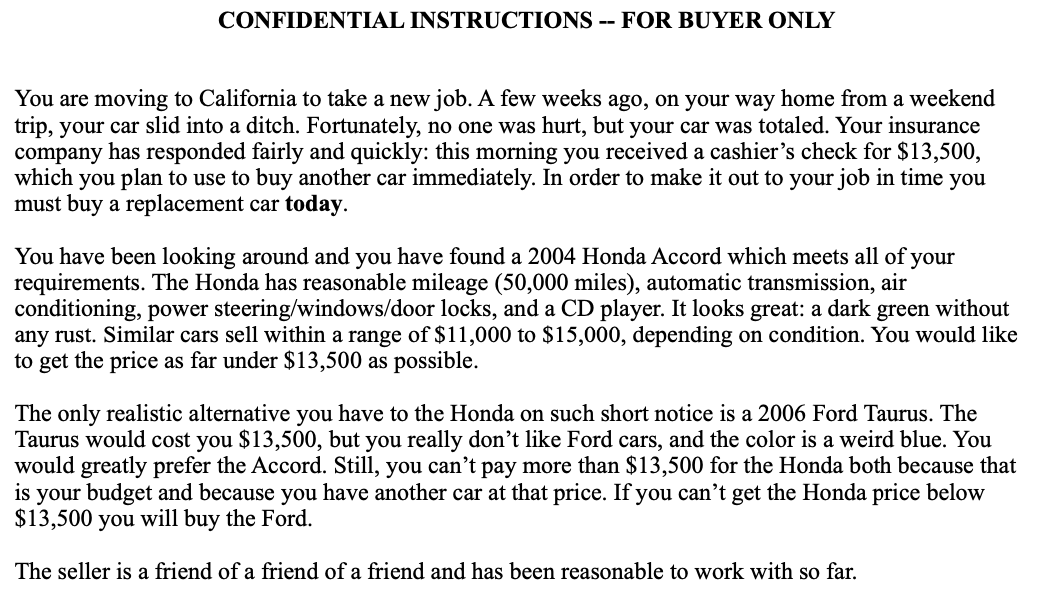}
    \caption{Honda scenario for the buyer.}
    \label{fig:honda_scenario_buyer}
\end{figure*}

\begin{figure*}[h]
    \centering
    \includegraphics[width=\textwidth]{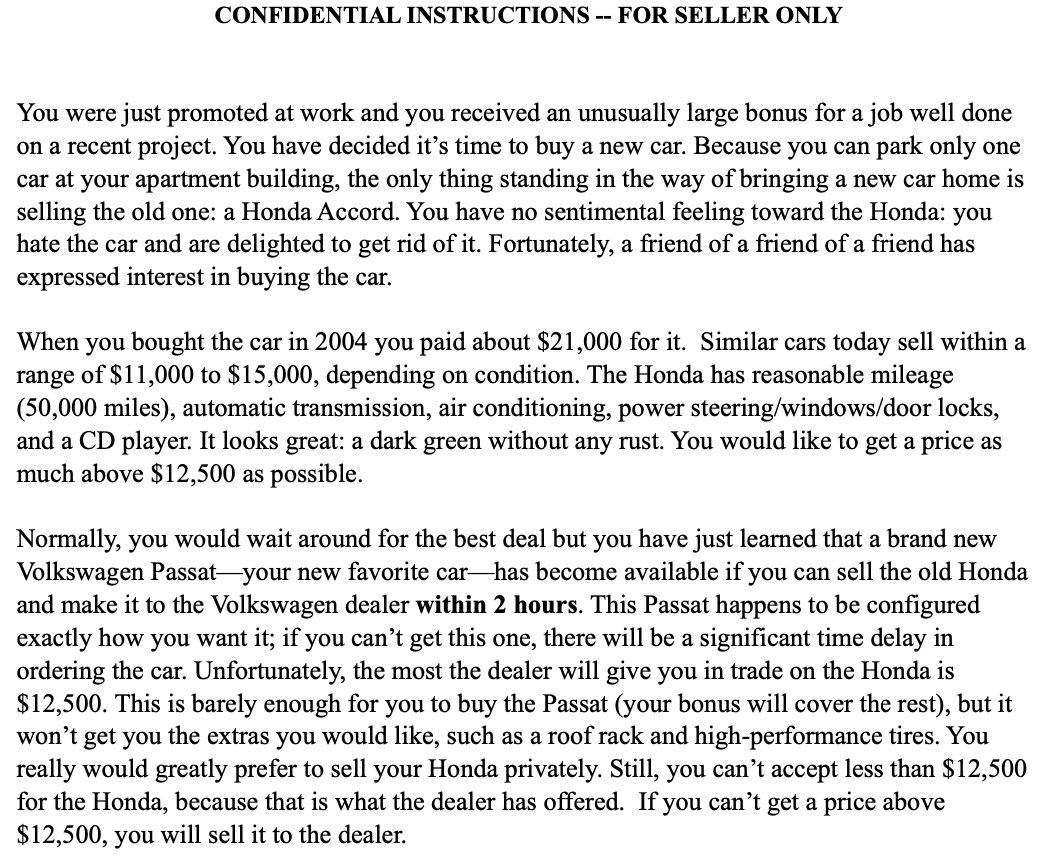}
    \caption{Honda scenario for the seller.}
    \label{fig:honda_scenario_seller}
\end{figure*}

\begin{figure*}[h]
    \centering
    \includegraphics[width=\textwidth]{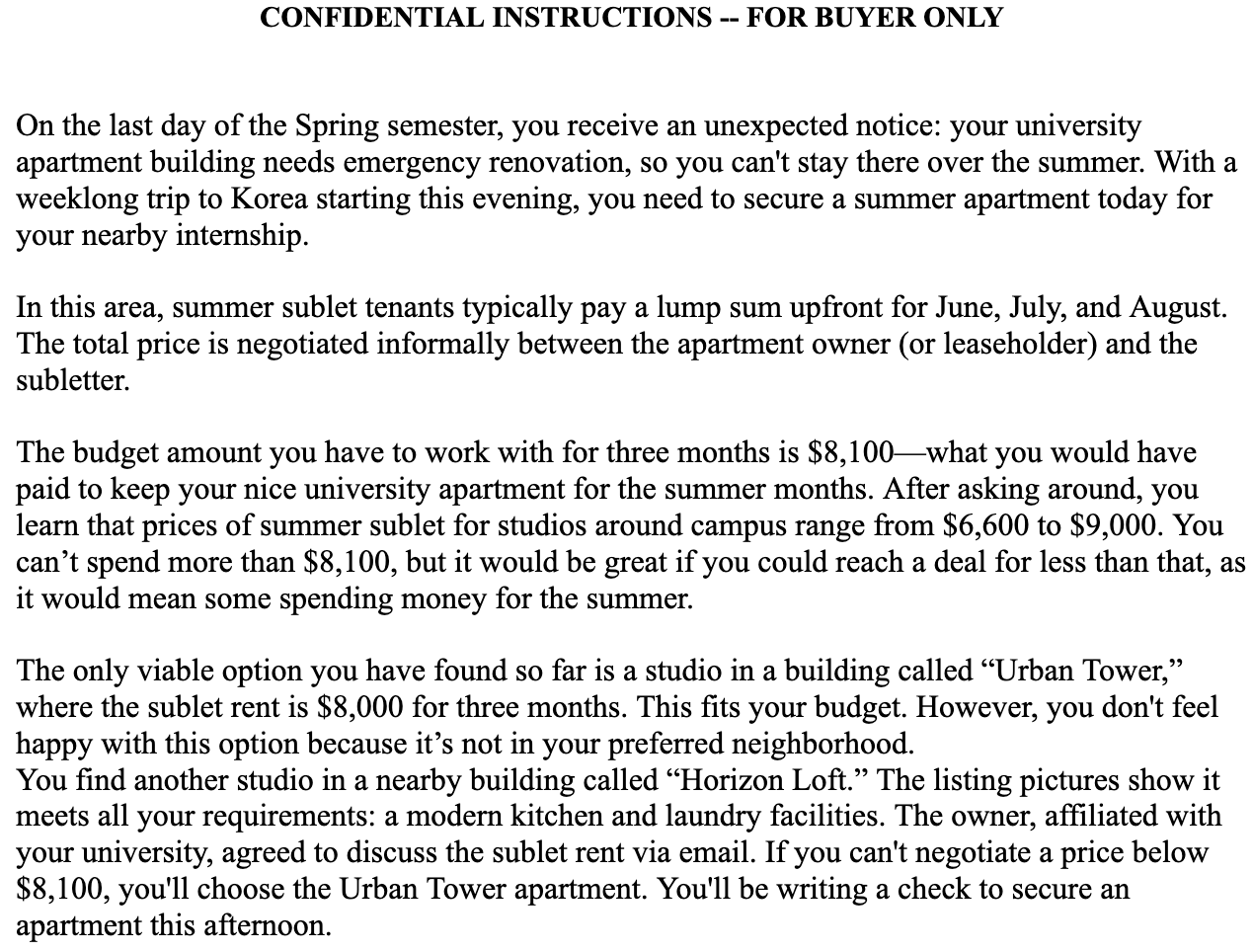}
    \caption{Summer sublease scenario for the buyer.}
    \label{fig:sublease_buyer}
\end{figure*}

\begin{figure*}[h]
    \centering
    \includegraphics[width=\textwidth]{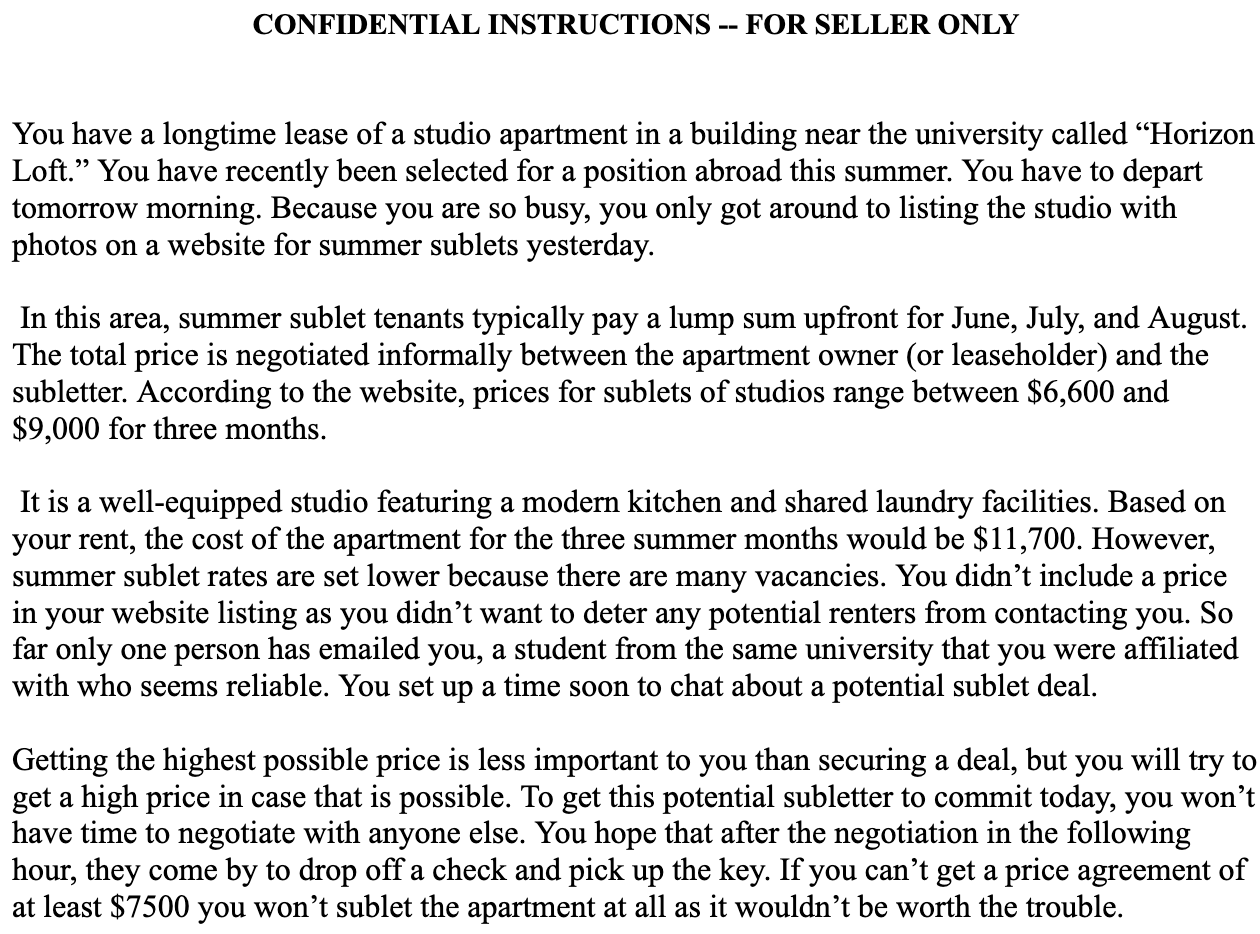}
    \caption{Summer sublease scenario for the seller.}
    \label{fig:sublease_seller}
\end{figure*}

Figures \ref{fig:honda_scenario_buyer} and \ref{fig:honda_scenario_seller} show the full negotiation scenarios for the first task in our collected dataset. These scenarios both relate to bargaining over a used Honda Accord. Figure \ref{fig:honda_scenario_buyer} shows the scenario given to the buyer and Figure \ref{fig:honda_scenario_seller} shows the scenario for the seller. These same scenarios are used in our user experiment along with the summer sublease scenarios in Figures \ref{fig:sublease_buyer} and \ref{fig:sublease_seller} (Section \ref{sec:evaluation}). The scenario for the buyer is given directly to the participants as part of the experiment instructions and the scenario for the seller is used to construct the system prompt for our negotiation agent. Table \ref{tab:chatbot_instructional_prompt} shows the instructional prompt for the used car scenario.

\section{Full Feedback}
\label{sec:full_feedback}

\begin{figure*}
    \centering
    \includegraphics[width=\textwidth]{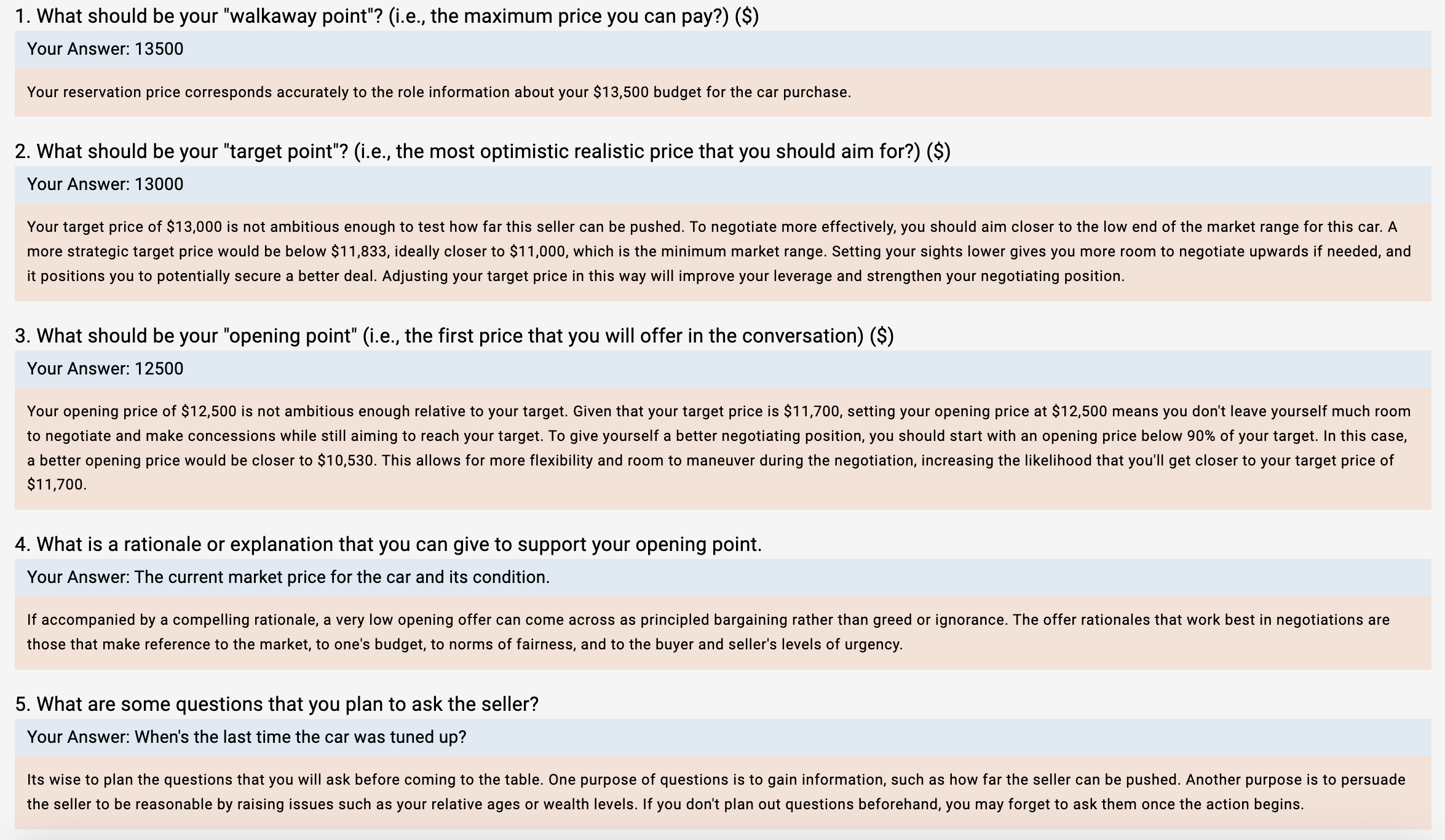}
    \caption{Full example of preparation feedback.}
    \label{fig:full_prep_feedback}
\end{figure*}

\begin{figure*}
    \centering
    \includegraphics[width=\textwidth]{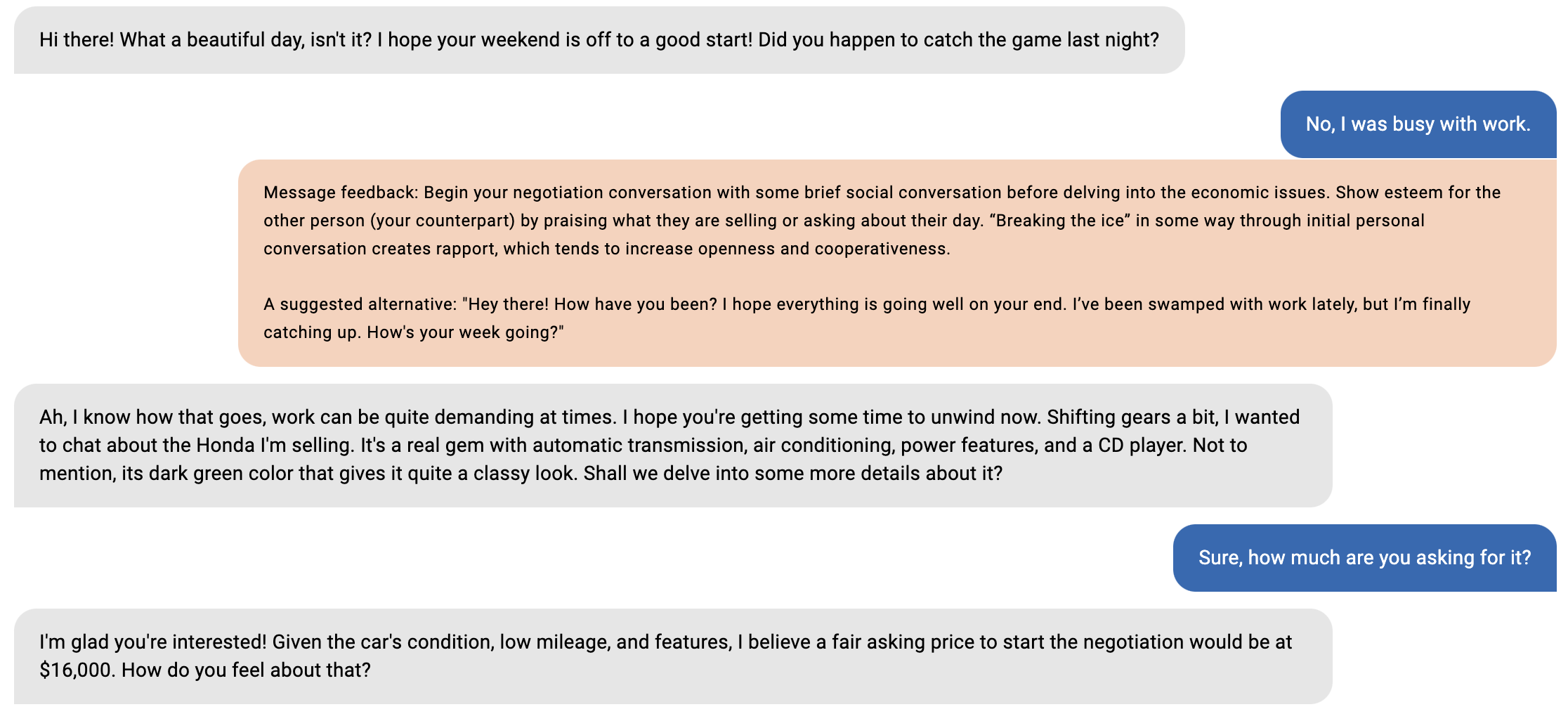}
    \caption{Example of turn based feedback where a user made an error in \textbf{breaking the ice}.}
    \label{fig:turn_based_feedback_icebreaker}
\end{figure*}

\begin{figure*}
    \centering
    \includegraphics[width=\textwidth]{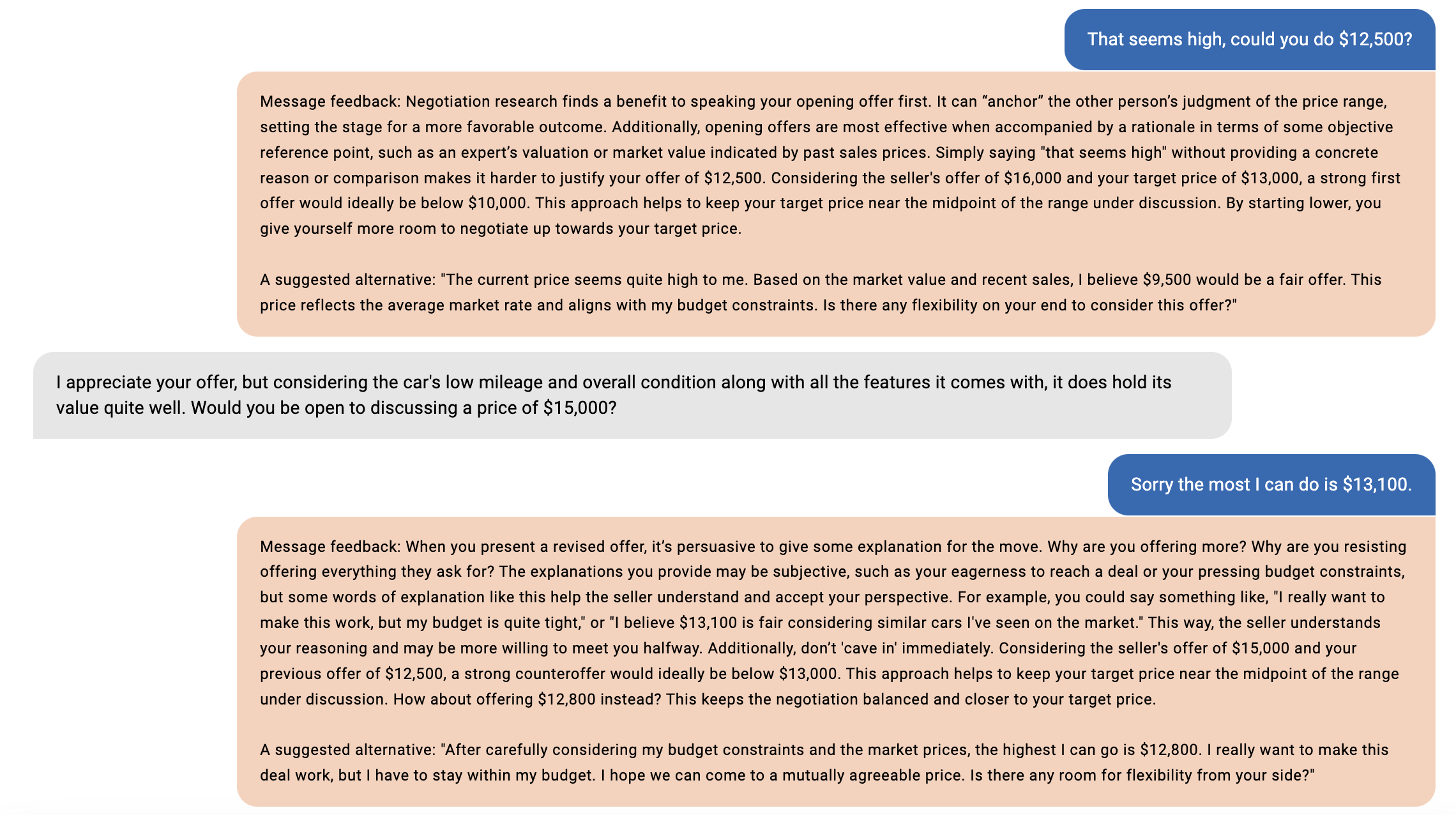}
    \caption{Example of two turn-based feedbacks from ACE. The first turn shows feedback where the user failed in \textbf{giving the first offer}, \textbf{ambitious opening point}, and \textbf{including rationale}. The second turn gives feedback where the user failed in \textbf{including rationale} and \textbf{strong counteroffer}.}
    \label{fig:turn_based_feedback_opener_counteroffer}
\end{figure*}

\begin{figure*}
    \centering
    \includegraphics[width=\textwidth]{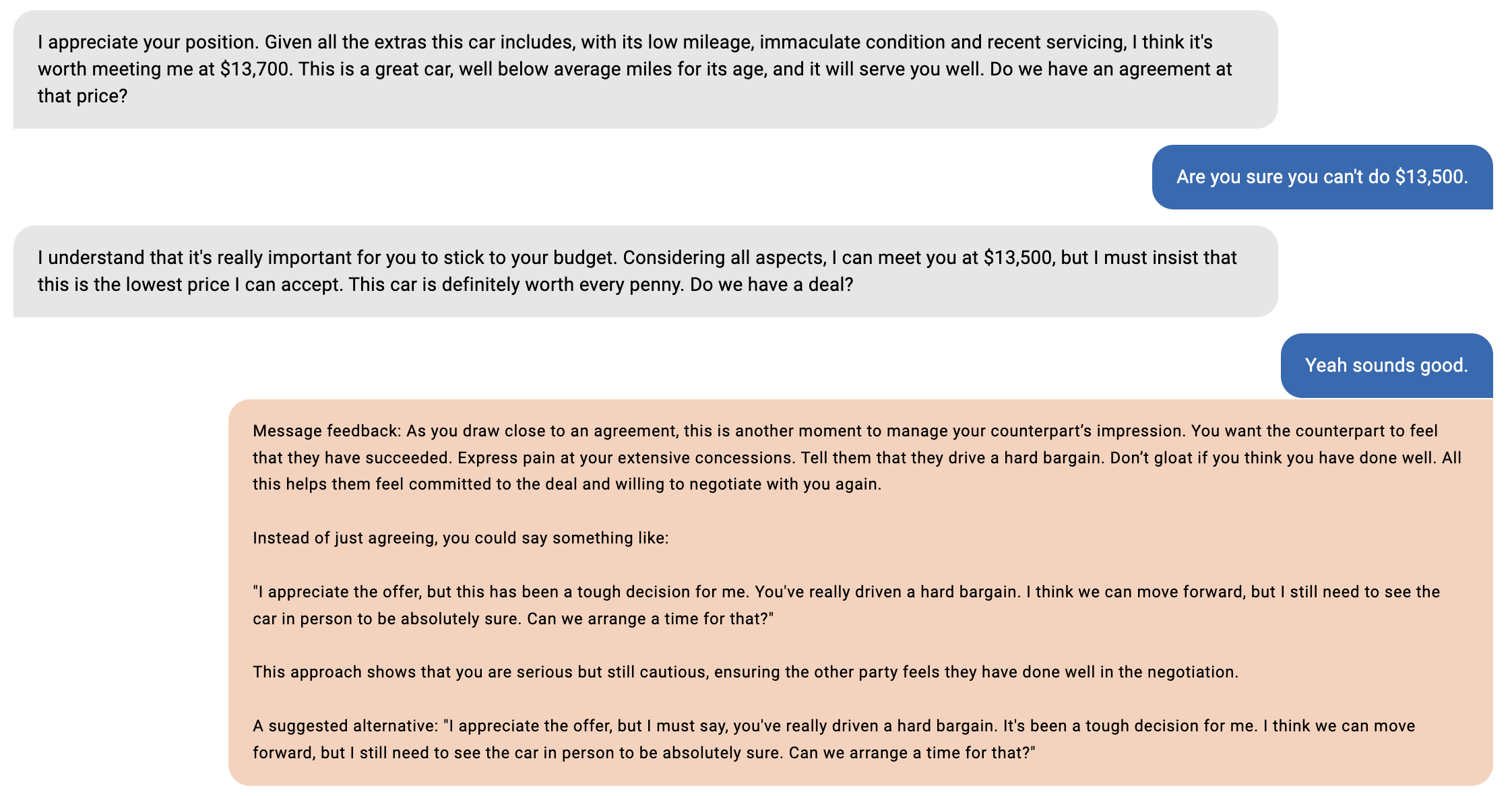}
    \caption{Example of turn based feedback where a user made an error in the \textbf{strategic closing} category.}
    \label{fig:turn_based_feedback_closer}
\end{figure*}

\begin{figure*}
    \centering
    \includegraphics[width=\textwidth]{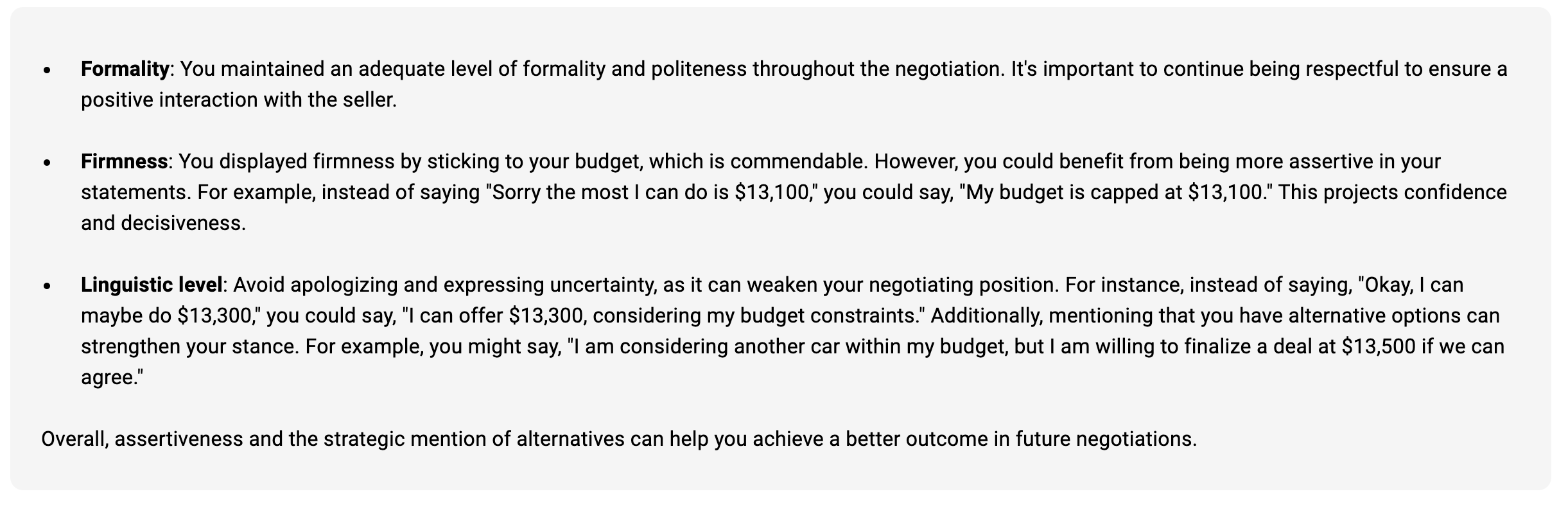}
    \caption{A complete example of holistic feedback from ACE.}
    \label{fig:full_overall_feedback}
\end{figure*}

Figures \ref{fig:full_prep_feedback} through \ref{fig:full_overall_feedback} show full examples of feedback given by ACE. All of the feedback is given according to the conversation in Table \ref{tab:convo_full} which is based on the Honda Accord negotiation scenario (Figures \ref{fig:honda_scenario_buyer} and \ref{fig:honda_scenario_seller}). Figure \ref{fig:full_prep_feedback} gives a complete example of preparation feedback, Figures \ref{fig:turn_based_feedback_icebreaker} to \ref{fig:turn_based_feedback_closer} show full examples of turn-based feedback, and Figure \ref{fig:full_overall_feedback} shows a complete example of holistic feedback.

\section{Prompts}
\label{sec:prompts}
Tables \ref{tab:price_extraction_prompt} through \ref{tab:chatbot_instructional_prompt} show the key prompts we use for building ACE. All prompts were engineered based on GPT-4 version GPT-4-0613 and GPT-4o version GPT-4o-2024-05-13. The includes prompts we use identify user errors (Tables \ref{tab:price_extraction_prompt} and \ref{tab:rationale_prompt}), giving direct feedback (Table \ref{tab:direct_feedback_prompts}), revising utterances (Table \ref{tab:re-edit_message_prompt}), preparation feedback (Table \ref{tab:target_price_prompts}), and holistic feedback (Table \ref{tab:overall_feedback_prompt}). We also include the prompt for our negotiation chatbot agent (Table \ref{tab:chatbot_instructional_prompt}).

\begin{figure*}[h]
    \centering
    \includegraphics[width=\textwidth]{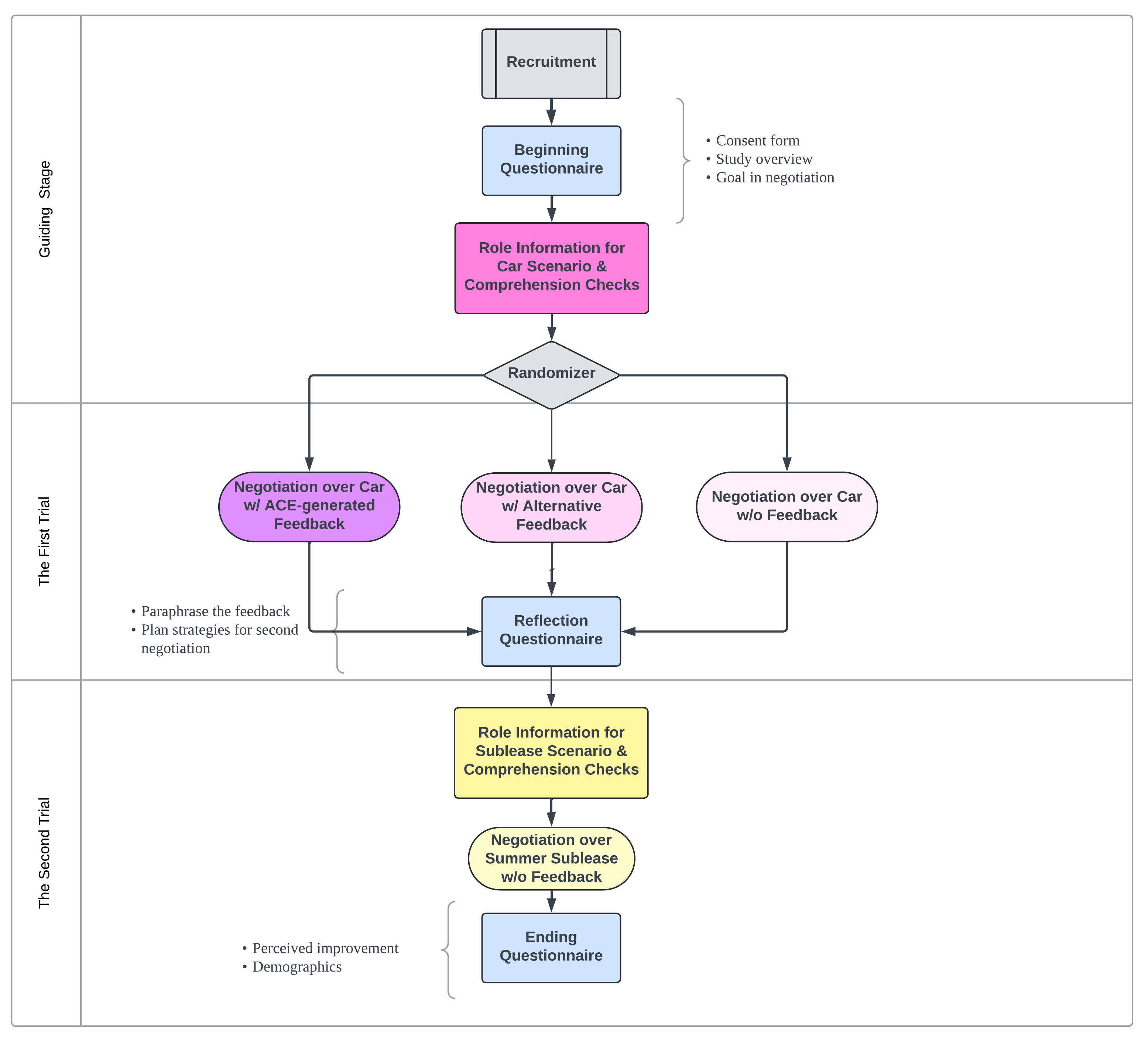}
    \caption{Experiment diagram, we designed our experiment in Qualtrics.}
    \label{fig:experiment_diagram}
\end{figure*}

\clearpage
\newpage

\begin{table*}
\centering
\begin{tabular}{p{5.6in}}
\hline \textbf{Example Conversation } \\ \hline
\textbf{AI Seller:} Hi there! What a beautiful day, isn't it? I hope your weekend is off to a good start! Did you happen to catch the game last night?
 \\
\textbf{User:} No, I was busy with work. \\
\textbf{AI Seller:} Ah, I know how that goes, work can be quite demanding at times. I hope you're getting some time to unwind now. Shifting gears a bit, I wanted to chat about the Honda I'm selling. It's a real gem with automatic transmission, air conditioning, power features, and a CD player. Not to mention, its dark green color that gives it quite a classy look. Shall we delve into some more details about it?\\
\textbf{User:} Sure, how much are you asking for it? \\
\textbf{AI Seller:} I'm glad you're interested! Given the car's condition, low mileage, and features, I believe a fair asking price to start the negotiation would be at \$16,000. How do you feel about that? \\
\textbf{User:} That seems high, could you do \$12,500?  \\
\textbf{AI Seller:} I appreciate your offer, but considering the car's low mileage and overall condition along with all the features it comes with, it does hold its value quite well. Would you be open to discussing a price of \$15,000? \\
\textbf{User:} Sorry the most I can do is \$13,100  \\
\textbf{AI Seller:} I appreciate your offer. Given the features and condition of the car, I can come down to \$14,500. It has a lot of value for that price considering its mileage and overall state. It's a solid, reliable car that still has a lot of life left in it. What do you think?\\
\textbf{User:} When's the last time you had it tuned up?  \\
\textbf{AI Seller:} I'm glad you asked! The car's maintenance has always been a top priority. It received its most recent complete tune-up just a couple of months ago. This included changing the engine oil, checking all fluids, replacing spark plugs, inspecting the brakes, and doing a thorough check of the entire vehicle. So it's in great running condition. Would this possibly influence your offer?\\
\textbf{User:} Okay, I can maybe do \$13,300.  \\
\textbf{AI Seller:} I understand where you're coming from and I do appreciate the offer. How about we meet halfway? I can lower the asking price to \$14,000. This car is a fantastic deal at that price and it would ensure you peace of mind for many ears to come given its condition and maintenance history. What do you say?\\
\textbf{User:} The most I can do is \$13,500  \\
\textbf{AI Seller:} I understand that it's really important for you to stick to your budget. Considering all aspects, I can meet you at \$13,500, but I must insist that this is the lowest price I can accept. This car is definitely worth every penny. Do we have a deal?\\
\textbf{User:} Yeah sounds good.  \\
\hline
\end{tabular}
\caption{An example conversation between a user and our LLM negotiation agent (Section \ref{sec:approach}).}
\label{tab:convo_full}
\end{table*}

\clearpage
\newpage

\begin{table*}
\centering
\begin{tabular}{p{5.6in}}
\hline \textbf{Price Extraction Prompt} \\ \hline
\textbf{Instructional Prompt:} \\
\#INSTRUCTION \\
You have to extract priced offers from messages. Just give the dollar amount and nothing else. 
If no offer was proposed yet then say so. If an offer was accepted then say so. If the offer
is presented as range of prices, then give both the prices. Do not ellicitate your reasoning. \\
\\
\#EXAMPLES \\
Message : "I will be willing to pay something from 10k to 11k" \\ 
Offer:  "10000 to 11000".
\\
\\
Message: "so i am uh looking for this car and my current price range is between uh eleven thousand and five hundred to twelve thousand dollars" \\
Offer: "11500 to 12000" 
\\
\\
Message: "Ooh, that's kind of rough. Our sticker price for this car is closer to \$14,000." \\
Offer: "14000"
\\
\\
Message: "Yes 12000 sounds like a good price for me." \\
Offer: "Accepted."
\\
\\
Message: "That's well beyond my price, I can't do that" \\
Offer: "Refused."
\\
\\
Message: "Sure. No Problem" \\
Offer: "No offer."
\\
\\
Message: "I don't think I am able to do that" \\
Offer: "Refused."
\\
\\
Message: "12,500... I mean, could we call it even \$13,000? \\
Offer: "13000"
\\
\\
Message: "You said you would be willing to pay 12k ?" \\
Offer: "Rephrasing."
\\
\\
\#EXTRACTION
\\
\\
Message: \{\}\\
Offer: \\
\hline
\end{tabular}
\caption{Prompt used to extract prices from user utterances. These prices are used to identify all of the price-based negotiation errors.}
\label{tab:price_extraction_prompt}
\end{table*}

\begin{table*}
\centering
\begin{tabular}{p{5.6in}}
\hline \textbf{Rationale Error Prompt} \\ \hline
\textbf{Instructional Prompt:} \\
\#INSTRUCTION \\
We are here in the context of a negotiation.
Your task is to detect if the buyer gives sufficient rationale/argument along with their offer or not in the passage.
\\
We think of rationales as a piece of argumentation that can justify a mentioned price. 
Rationales should be related to the item purchased (clearly mentioning some specific features or price ranges etc…). 
Nevertheless, we can think of exceptions such as “You're a friend so I can maybe push it a bit to…” \\
\\
\#EXAMPLE \\
\\
Passage: "Buyer: Hello I'd like to make an offer\\
              Seller: Great what were you thinking ?\\
              Buyer: I don't know something like 10k ?"\\ 
Rationale :False
\\
\\
Passage: "Buyer: Hello, this car is in great shape for its mileage, I was looking
for a similar car on the internet. I like it and my kids would have a great time in it.
Can I make an offer ? \\
Seller: Sure how much ? \\
Buyer: Something around 10k ?"\\
Rationale :True
\\
\\
Passage: "Buyer: Yeah I guess i can do 12,500. It seems reasonable. \\
Seller: Can you push it more to 13,5? \\
Buyer: No sorry, 12,5 nothing more."\\
Rationale : False
\\
\\

\#Task
\\
\\
Passage : \{\}\\
Rationale :Answer here\\
\hline
\end{tabular}
\caption{Prompt used to identify errors related to \textbf{including rationale}.}
\label{tab:rationale_prompt}
\end{table*}

\begin{table*}
\centering
\begin{tabular}{p{5.6in}}
\hline \textbf{Direct Feedback Prompts} \\ \hline
\textbf{Rationale Feedback Prompt:} \\
\#INSTRUCTION \\
We are here in the context of a negotiation. You are an assistant aimed help a buyer in a negotiation and give them comments on their offers. In this passage: \{\} \\
\\
The buyer did not give enough arguments to justify their offer.\\
\\
Give the buyer a quick explanation. Try to quote some words the buyer said.\\
\\
EXAMPLE OF EXPLANATION: \\
"When you present a revised offer, it’s persuasive to give some explanation for the move. Why are you offering more? Why are you resisting offering everything they ask for? The explanations you provide may be subjective, such as your eagerness to reach a deal or your pressing budget constraints, but some words of explanation like this help the seller understand and accept your perspective. "
\\
\\
\textbf{Counteroffer Feedback Prompt:}\\
You are an assistant aimed to reedit text to help a buyer in a negotiation and provide them feedback on their offer.\\
Here is the conversation :\\
\{\} \\
\\
Give them an explanation.\\
\\
Example of good explanation:\\
\\
"Considering the seller's offer of \$\{\} and your target price of \$\{\}, a strong first offer would ideally be below \$\{\}. This approach helps to keep your target price near the midpoint of the range under discussion."\\
\hline
\end{tabular}
\caption{Prompt used to give direct feedback to users.}
\label{tab:direct_feedback_prompts}
\end{table*}

\begin{table*}
\centering
\begin{tabular}{p{5.6in}}
\hline \textbf{Re-edit Message Prompt} \\ \hline
\textbf{Instructional Prompt:} \\
We are in the context of a negotiation.
Different teachers gave comments to the buyer: \\
Your task is to propose an alternative message the buyer could have sent that would match all the comments given by teachers. \\
\\
For example if a comment is saying that the buyer should open the conversation with an ice breaker, then propose an icebreaker.
If a comment is saying that they should add rationales to their offers, then rewrite the offer and add a few rationales to it.
You have to put yourself in the buyer's position. Assume that you are talking to the seller. \\
\\
\#EXAMPLE1 :\\
- MESSAGE:\\
"Seems a little steep, steep for me. You know, I can do something in the, you know, \$12,000 range would really be, you know, near the top of the end of my budget. Do you have any flexibility there? You know, anything we can do to, you know, work on that price?"
\\
\\
-COMMENTS:
"comment 1: "Negotiation research finds a benefit to speaking your opening offer first. It can “anchor” the other person’s judgment of the price range, setting the stage for a more favorable outcome." \\
comment 2: "Considering your target price of \$10000, a strong first offer would ideally be below \$9000. This approach helps to keep your target price near the midpoint of the range under discussion."
\\
\\
- ANSWER: "The price seems a little steep for me. I can work with something in the \$9,000 range, which is near the top end of my budget. I want to ensure that we can reach a mutually beneficial agreement. Is there any flexibility on the price from your end?"
\\
\\
\#EXAMPLE2:\\
-MESSAGE:\\
"Hi, I'm looking for probably a Honda Accord with reasonable mileage around maybe \$15000. Do you have anything like that?"
\\
\\
-COMMENTS:\\
"comment 1: "Begin your negotiation conversation with some brief social conversation before delving into the economic issues. Show esteem for the other person (your counterpart) by praising what they are selling or asking about their day. “Breaking the ice” in some way through initial personal conversation creates rapport, which tends to increase openness and cooperativeness.\\
comment 2: "Negotiation research finds that opening offers are most effective when accompanied by a rationale in terms of some objective reference point, such as an expert’s valuation of the object under negotiation or market value indicated by past sales prices."\\
-ANSWER: "Hey ! It has been a long time are you doing ?"
\\
\\
\#YOUR TURN TO DO IT NOW \\
-MESSAGE: \\
\{\}\\
- COMMENTS:\\
\{\}\\
- ANSWER:\\
\hline
\end{tabular}
\caption{Prompt used to revise user messages.}
\label{tab:re-edit_message_prompt}
\end{table*}

\begin{table*}
\centering
\begin{tabular}{p{5.6in}}
\hline \textbf{Relevant Target Price Prompts} \\ \hline
\textbf{Low Target Feedback Prompt:} \\
You are an assistant aimed to give advice to help a buyer in a negotiation. You are addressing directly to the buyer, use the second person (You). \\
\\
The buyer made an error setting their target price for the negotiation. The buyer set their target price to \$\{\}. However a good target price should be above the minimum market value for the car which is \$\{\}.\\
\\
Give the buyer feedback explaining their error including details about what would be a good target price. \\
Here is an example of good feedback: \\
This overly ambitious target is below the market range for the car.  It may cause offense. By overreaching, you may miss out on good deal.\\
\\
\textbf{High Target Feedback Prompt:} \\
You are an assistant aimed to give advice to help a buyer in a negotiation. You are addressing directly to the buyer, use the second person (You).
\\
\\
The buyer made an error setting their target price for the negotiation. The buyer set their target price to \$\{\}. However a good target price should be below \$\{\} and closer to the minimum market range for the car which is \$\{\}.\\
\\
Give the buyer feedback explaining their error including details about what would be a good target price. \\
Here is an example of good feedback: \\
Your target price of \{\} is not ambitious enough to test how far this seller can be pushed. You should aspire to a price at the low end of the market range.\\
\hline
\end{tabular}
\caption{Prompt used to give feedback on how well the user prepared their target price.}
\label{tab:target_price_prompts}
\end{table*}

\begin{table*}
\centering
\begin{tabular}{p{5.6in}}
\hline \textbf{Holistic Feedback Prompt} \\ \hline
\textbf{Low Target Feedback Prompt:} \\
Given the negotiation transcript: \{\} \\
\\
Your goal is to to build a constructive feedback to a user in order to them reaching a better outcome if they had to go over this
negotiation again. You will focus on the linguistics aspect and strategic aspects and dont bother with discussing the prices offered.
You are adressing directly to the buyer, use the second person (You).
Here are the dimensions your feedback will include:\\
\\
- Formality: A buyer cannot be rude and pushy. Also a good buyer stays polite. \\
- Firmness: A buyer cannot be too emotional. Studied have shown that firm and tough levels of communication help reaching better economic outcome than warmth and too friendly.\\
- Linguistic level: A buyer should not be apologizing. Buyer do not say the word “greedy” (can be interpreted as a personal attack).\\
As a buyer you should project that you do not need to buy a car/you have a perfectly good alternative. The buyer also should somehow mention that they have a plan B.\\
\\
Feedback:\\
\hline
\end{tabular}
\caption{Prompt used to give holistic feedback to the user.}
\label{tab:overall_feedback_prompt}
\end{table*}

\begin{table*}
\centering
\begin{tabular}{p{5.6in}}
\hline \textbf{Negotiation Agent Prompt} \\ \hline
\textbf{Instructional Prompt:} \\
You are a chatbot designed for negotiation. The discussion has to be fluent and realistic.
The Honda has reasonable mileage (50,000 miles), automatic transmission, air conditioning,
power steering/windows/door locks, and a CD player. It looks great: a dark green without any rust.
You need to sell the car for a price above \$\{\}. You will not sell the car for below that amount. You are selling the car as a private individual not a dealer.
Try not to be redundant in your arguments and talks (do not repeat what you already said in previous turns). If you give the buyer a counteroffer make sure any new offers are lower than the price you gave previously. Make sure you negotiate hard and never offer a price lower than what the buyer gives you.
Do not mention that you need to sell the car for over \$\{\}. If the buyer offers a price below \$8,000 respond with "That's a very unrealistic price.  Please start with an offer that aligns with the market range for this kind of car.  Otherwise I can't take time to talk with you about this car." \\
\hline
\end{tabular}
\caption{The instructional prompt for our negotiation chatbot agent.}
\label{tab:chatbot_instructional_prompt}
\end{table*}

\end{document}